\documentclass[preprint]{agujournal}
\usepackage{amsfonts,amssymb}
\usepackage{amsmath}
\usepackage{float}
\usepackage{bm}
\usepackage{url}
\usepackage{algorithm}
\usepackage[noend]{algpseudocode}
\usepackage{color}
\usepackage{booktabs, caption, makecell}
\usepackage{threeparttable}
\usepackage{multirow}
\usepackage{multicol}

\journalname{Water Resources Research}

\begin{document}

\title{Deep autoregressive neural networks for high-dimensional inverse problems in groundwater contaminant source identification}
\justify
\authors{Shaoxing Mo\affil{1,2}, Nicholas Zabaras\affil{2}, Xiaoqing Shi\affil{1}, and Jichun Wu\affil{1}}

\affiliation{1}{Key Laboratory of Surficial Geochemistry of Ministry of Education, School of Earth Sciences and Engineering, Nanjing University, Nanjing, China.}

\affiliation{2}{Center for Informatics and Computational Science, University of Notre Dame, Notre Dame, IN, USA.}
 
\correspondingauthor{Nicholas Zabaras}{nzabaras@gmail.com}
\correspondingauthor{Xiaoqing Shi}{shixq@nju.edu.cn}

\begin{keypoints}
\item A deep neural network surrogate approach is developed for groundwater contaminant transport in highly-heterogeneous media
 \item The time-varying process is captured using an autoregressive model 
 \item A $686$-dimensional inverse problem of contaminant source identification in heterogeneous media is addressed
\end{keypoints}

\begin{abstract}
 Identification of a groundwater contaminant source simultaneously with the hydraulic conductivity in highly-heterogeneous media   often results in a high-dimensional inverse problem. In this study, a deep autoregressive neural network-based surrogate method is developed for the forward model to allow us to solve efficiently such high-dimensional inverse problems. The surrogate is trained using limited evaluations of the forward model. Since the relationship between the  time-varying inputs and outputs of the forward transport model is complex, we propose an autoregressive  strategy, which treats the output at the previous time step as input to the network for predicting the output at the current time step. We employ a dense convolutional encoder-decoder network architecture in which the high-dimensional input and output fields of the model are treated as images to leverage the robust capability of convolutional networks in image-like data processing. An iterative local updating ensemble smoother (ILUES) algorithm is used as the inversion framework. The proposed method is evaluated using a synthetic contaminant source identification problem with $686$ uncertain input parameters. Results indicate that, with relatively limited training data, the deep autoregressive neural network consisting of $27$ convolutional layers is capable of providing an accurate approximation for the high-dimensional model input-output relationship. The autoregressive strategy substantially improves the network's accuracy and computational efficiency. The application of the surrogate-based ILUES in solving the inverse problem shows that it can achieve accurate inversion results and predictive uncertainty estimates.
 
 \noindent\textbf{Keywords:} Inverse problem; source identification; high-dimensionality; deep autoregressive neural networks; iterative ensemble smoother

\end{abstract}

\section*{Plain Language Summary }
High-dimensional inverse problems are often computationally expensive since a large number of forward model evaluations are usually required. A computationally efficient alternative is to replace in the inversion process the forward model with an accurate but fast-to-evaluate surrogate model. However, most existing surrogate methods suffer from the ``curse of dimensionality". In this paper, we develop a deep autoregressive neural network-based surrogate method to efficiently solve such high-dimensional inverse problems. The autoregressive neural network can efficiently obtain accurate approximations for the time-dependent outputs of forward models with time-varying inputs. In addition, a dense convolutional network architecture is employed to transform the surrogate modeling task to an image-to-image regression problem by leveraging the robust capability of convolutional networks in image-like data processing. The curse of dimensionality was tackled though a series of inherent nonlinear projections of the input into low-dimensional latent spaces. It is shown that the proposed network provides an accurate surrogate of a transport system with high-dimensional uncertain inputs using relatively limited training data. Reliable inversion results are then obtained with a rather minor computational cost by using the surrogate to replace the forward model in the inversion process.

\section{Introduction}
 Identification of groundwater contaminant sources (including the location and release history)  together with the estimation of the  highly-heterogeneous aquifer hydraulic conductivity field are crucial for accurate prediction of the effects of contamination and for subsequent science-informed remedy decision making~\citep{DATTA2009,zhang2016,xu2018}. In practice, direct identification of the contaminant source and conductivity field is unfeasible. Therefore, they are commonly identified by solving an inverse problem based on easily accessible measurements of the state variables such as the contaminant concentrations and hydraulic heads~\citep{xu2018,zhang2015,zhang2016}. However, due to the strongly-heterogeneous nature of the aquifer conductivity field, even if parameterized using some dimensionality reduction methods (e.g., Karhunen-Lo{\` e}ve expansion (KLE), see~\citet{zhang2004}), it often needs a rather large number of stochastic degrees of freedom for accurate representation of the heterogeneity. As a consequence, the joint identification of contaminant source and conductivity field often leads to a high-dimensional inverse problem.
 
 Given the collected measurements, various approaches are available to solve the inverse problem~\citep{Rajabi2018,zhou2014}. The  Bayesian inference based on Markov chain Monte Carlo  sampling~\citep{Haario2006,vrugt2016} treats the model input parameters and outputs as random variables to formulate the inverse problem in a probabilistic framework~\citep{vrugt2016,zhang2013,zhang2015,zhang2016}. However, its applicability towards high-dimensional problems is computationally prohibitive as a large number of model evaluations are often required to obtain converged solutions~\citep{Chaudhuri2018,Keller2018,zhang2018}. As relatively faster alternatives to the Markov chain Monte Carlo methods, ensemble-based data assimilation algorithms such as ensemble Kalman filter (EnKF)~\citep{Evensen1994} and ensemble smoother (ES)~\citep{VanLeeuwen1996} have been widely used in groundwater hydrology to solve the inverse problem~\citep [e.g.,][]{Chang2017,Chaudhuri2018,Chen2006,Elsheikh2013,Emerick2012,Emerick2013,ju2018,Keller2018,li2012,sun2009a,sun2009b,White2018,xia2018,xu2018,zhang2018,zhou2011}. Similar to the Markov chain Monte Carlo methods, the ensemble-based data assimilation algorithms can provide simultaneously estimations of the parameter values and of the associated uncertainty, which enables further uncertainty analysis on predictions.

 Compared to EnKF which needs to update the model parameters and states simultaneously, ES updates the parameters solely, avoiding the inconsistency issue between the updated parameters and states observed in EnKF~\citep{Chang2017,ju2018,Thulin2007,zhang2018}. Many iterative variants of ES have been developed to improve its applicability towards strongly nonlinear and high-dimensional problems~\citep [e.g.,][]{Chen2012,Chen2013,Emerick2013,li2018,Stordal2015,zhang2018}, among which the iterative local updating ensemble smoother (ILUES) proposed in~\citet{zhang2018} shows a promising performance. In ILUES, rather than updating the ensemble of uncertain parameters globally, each sample in the ensemble is locally updated using its neighboring samples~\citep{zhang2018}. In addition, the observation data are assimilated multiple times with an inflated covariance matrix of the measurement errors to improve the data match quality~\citep{Chang2017,Emerick2013,zhang2018}. In this study, the ILUES algorithm is employed as the inversion framework to solve high-dimensional inverse problems.
 
Like other Monte Carlo-based methods, ILUES suffers from the curse of dimensionality. When the number of uncertain inputs is large and the input-output relationship is complex, ILUES requires a large ensemble size and iteration number to guarantee reliable estimations of parameter values as well as parametric uncertainty~\citep{zhang2018}, leading to a large computational burden. One effective solution to reduce the computational cost is to use surrogate models which can provide an accurate but fast-to-evaluate approximation of the model input-output relationship~\citep{Asher2015,Razavi2012}. Many surrogate methods based on, e.g., the polynomial chaos expansion (PCE)~\citep{xiu2002} and Gaussian processes~\citep{Williams2006}, have been applied widely to address  inverse problems in groundwater hydrology. They have shown an impressive approximation accuracy and computational efficiency in comparison to traditional inversion methods~\citep{Chang2017,Elsheikh2014,ju2018,laloy2013,ma2009,zeng2010,Zeng2012,zhang2015,zhang2016}. Two recent applications of the surrogate-based iterative ES in estimation of the heterogeneous conductivity field in subsurface flow problems can be found in~\citet{Chang2017} and~\citet{ju2018}, where~\citet{Chang2017} used two surrogate methods including a polynomial chaos expansion and a stochastic collocation method, and~\citet{ju2018} employed the Gaussian process for surrogate construction.  
 
 Unfortunately, most existing surrogate methods also suffer from the curse of dimensionality due to the exponential increase in the computational cost for surrogate construction as the input dimenionality increases~\citep{Asher2015,lin2009,liao2017,mo2018,Razavi2012}. If the output responses are strongly nonlinear, the computational cost will be further increased as more training samples are required to obtain desired approximation accuracy. As a result, in previous studies the surrogate methods were applied either in relatively low-dimensional ($<110$) problems or relatively simple flow models~\citep [e.g.,][]{laloy2013,ju2018,liao2013,liao2017,lin2009,ma2009,Zeng2012,zhang2015,zhang2016}. For example,~\citet{laloy2013} and~\citet{Chang2017} used PCE to construct surrogate models for groundwater flow models with $102$ and $55$ uncertain parameters, respectively.~\citet{ju2018} and~\citet{zhang2016} employed Gaussian process to build surrogate models for a groundwater flow model with $100$ uncertain parameters and a more nonlinear groundwater solute transport model with $13$ uncertain parameters, respectively. One approach to alleviate the curse of dimensionality is to use adaptivity for selecting informative training samples for surrogate construction~\citep [e.g.,][] {mo2017,zhang2013}. Such adaptive strategies can somewhat reduce the number of training samples, but the improvement is relatively limited for surrogate construction in problems of high-input dimensionality. 
 
 Recently, deep neural networks (DNNs) have gained an increased popularity in many communities due to their robustness and generalization property~\citep{laloy2017,laloy2018,Marcais2017,mo2018,shen2018,shen-etal2018,sun2018,zhu2018}. Compared to traditional surrogate methods, the large-depth architectures substantially increase the network's capability in representing complex functions, capturing abstract spatiotemporal features hidden in data, and learning advanced representations~\citep{shen2018}. For surrogate modeling tasks, the DNNs can provide accurate approximations for complex functions through a hierarchy of hidden layers of increasing complexity. They also scale well with the dimensionality due to the minibatch approach used when training the networks~\citep{Goodfellow-et-al-2016}.
 
 Three applications of DNNs for surrogate modeling tasks in uncertainty quantification for groundwater models were reported recently in~\citet{Tripathy2018} and our previous works as presented in~\citet{zhu2018} and~\citet{mo2018}. In~\citet{Tripathy2018}, a fully-connected DNN was used for surrogate modeling and the network was able to make predictions given $1024$-dimensional input conductivity fields at arbitrary correlation lengths. In our previous works~\citep{mo2018,zhu2018}, we proposed a dense convolutional encoder-decoder network architecture. The surrogate modeling task with high-dimensional model input and output fields was transformed to an image-to-image regression problem to leverage the robust capability of convolutional networks in image-like data processing~\citep{laloy2017,laloy2018,shen2018}. The dense convolutional network architecture, which introduces connections between non-adjacent layers to enhance the information flow through the network, was employed to alleviate the data-intensive  DNN training. The curse of dimensionality was tackled though a series of inherent nonlinear projections of the high-dimensional input into exploitable low-dimensional latent spaces.  This network architecture showed a promising computational efficiency in obtaining accurate surrogates to a single-phase flow model and a strongly nonlinear multiphase flow model with $4225$ and $2500$ input dimensions, respectively.

 In this paper, we adopt the dense convolutional encoder-decoder network architecture employed in our previous studies~\citep{mo2018,zhu2018} to formulate a deep autoregressive neural network based ILUES method to efficiently solve the high-dimensional inverse problem in groundwater contaminant source identification. As will be shown in this study, direct implementation of this network architecture to predict the time-dependent output fields of the contaminant transport model with a time-varying source term leads to inaccurate approximations especially when the available training data are limited. To improve the approximation accuracy and computational efficiency, we propose to use an autoregressive network that treats the system output at the previous time step as input to the network to predict the outputs at the current time step. Additional inputs to this autoregressive network include the time-dependent source term in the previous time interval, and the time-independent conductivity field. We will show that this autoregressive strategy can substantially improve the accuracy and efficiency of the network. In addition, to improve the accuracy in approximating the strongly nonlinear (large gradient) nature of the concentration in the region near the source release location due to small dispersivity, we consider in the loss function an additional term  in terms of  the concentrations in this region. The overall integrated methodology is applied to a synthetic groundwater contaminant source identification  problem with $686$ unknown input parameters, in which the contaminant source and highly-heterogeneous conductivity field are jointly estimated. 
 
 The rest of the paper is organized as follows. In section~\ref{sec:GovE}, we introduce a solute transport model and define the inverse problem of interest with a high-dimensional input. The ILUES algorithm and the deep autoregressive network based on a dense convolutional encoder-decoder network architecture are presented in section~\ref{sec:method}. In sections~\ref{sec:application} and~\ref{sec:results}, we evaluate the performance of the method using a synthetic model. Finally, the conclusions together with potential extensions are summarized in section~\ref{sec:conclusion}.
 
\section{Governing Equations and Problem Formulation}\label{sec:GovE}
 In this study, we consider a contaminant transport system under  steady-state groundwater flow conditions and assume that only advection and dispersion are considered as  transport mechanisms. The governing equation of the steady-state groundwater flow can be written as follows:
 \begin{linenomath*}
 \begin{equation}
     \nabla\cdot(K\nabla h)=0,
 \label{eq:flow}
 \end{equation}
 \end{linenomath*}
 and the pore space flow velocity $v\;[\rm{LT}^{-1}]$ can be obtained by  Darcy's law:
 \begin{linenomath*}
 \begin{equation}
     \bm v=-\frac{K}{\phi}\nabla h,
 \label{eq:darcy-flow}
 \end{equation}
 \end{linenomath*}
 where $K\;[\rm{LT}^{-1}]$ is the hydraulic conductivity, $\phi$ [-] is the effective porosity, and $h\;[\rm{L}]$ is the hydraulic head. The flow equation is numerically solved using the groundwater flow simulator MODFLOW~\citep{Harbaugh2000}. The resulting velocity $\bm v$ is then used as the input in the following advection-dispersion equation to calculate the contaminant concentration $c\;[\rm{ML}^{-3}]$~\citep{zheng1999}:
 \begin{linenomath*}
 \begin{equation}
     \frac{\partial(\phi c)}{\partial t}=\nabla\cdot(\phi \bm{\alpha}\nabla c)-\nabla\cdot(\phi \bm vc)+q_sc_s,
 \label{eq:adv-disp}
 \end{equation}
 \end{linenomath*}
 where $t$ [T] is time, $\bm\alpha\;[\rm{L}^2\rm{T}^{-1}]$ is the dispersion tensor determined by $\bm v$, and longitudinal ($\alpha_L$) and transverse ($\alpha_T$) dispersivities~\citep{zheng1999}, $q_s\;[\rm{T}^{-1}]$ is the volumetric flow rate per unit volume of aquifer representing fluid sources or sinks, and $c_s\;[\rm{ML}^{-3}]$ is the concentration of the source or sink. The solute  transport simulator MT3DMS~\citep{zheng1999} is used to solve the contaminant transport equation. 
 
 We are concerned with a groundwater contaminant source identification problem in highly-heterogeneous media. This involves identifying parameters that characterize the contaminant source, such as the source location and release strength  and   the highly-heterogeneous aquifer conductivity field. We pose this problem as   an inverse problem using   indirect measurements that include the contaminant concentration and hydraulic head. More specifically, we assume that the source strength is time-varying and can be characterized by a vector $\bm{S}_s\in\mathbb{R}^{n_t}$, where $S_{sj},\;j=1,\,\ldots,\,n_t,$ denotes the strength in the $j$-th time interval. The conductivity field $\mathbf{K}$ is treated as a random field. Therefore, along with the source location $\bm{S}_l$, the objective of this study is to simultaneously identify the contaminant source (i.e., $\bm{S}_l$ and $\bm{S}_s$) and the heterogeneous conductivity field $\mathbf{K}$. This inverse problem is often high-dimensional with a large number of unknown parameters as a large number of stochastic degrees of freedom is often required to accurately characterize the conductivity heterogeneity.

\section{Methods}\label{sec:method}

\subsection{Iterative Local Updating Ensemble Smoother}
 A common stochastic representation of a hydrological system can be written as
 \begin{linenomath*}
 \begin{equation}
     \bm{d}=f(\bm{m})+\bm{e},
 \end{equation}
 \end{linenomath*}
 where $\bm m$ is a $N_m\times 1$ vector of the uncertain input parameters, $\bm{d}$ is a $N_d\times 1$ vector of the measurements, $f(\cdot)$ denotes the nonlinear forward model with input $\bm{m}$, and $\bm{e}$ is a vector of the measurement errors. 
 
In ILUES~\citep{zhang2018}, the observation data are iteratively assimilated for multiple times. Rather than updating the ensemble of parameter realizations globally, ILUES employs a local update scheme to better handle strongly nonlinear and high-dimensional problems. More specifically, for an ensemble of $N_e$ parameter samples $\mathbf M=[\bm m_1,\ldots,\bm m_{N_e}]$, it first identifies a local ensemble for each sample $\bm m_i\in\mathbf M$ based on a combined measure of the normalized distance to the measurements $\bm d$ and sample $\bm m_i$:
 \begin{linenomath*}
 \begin{equation}\label{eq:J-value}
     J(\bm m)=\frac{J_d(\bm m)}{J_d^{\text{max}}} + \frac{J_m(\bm m)}{J_m^{\text{max}}},
 \end{equation}
 \end{linenomath*}
 where $J_d(\bm m)=[f(\bm m)-\bm d]^{\top}\mathbf C_{\text D}^{-1}[f(\bm m)-\bm d]$ is the distance between the model responses $f(\bm m)$ and measurements $\bm d$, and $J_m(\bm m)=(\bm m-\bm m_i)^{\top}\mathbf C_{\text{MM}}^{-1}(\bm m-\bm m_i)$ is the distance between the sample $\bm m_i$ and sample $\bm m\in\mathbf M$. Here, $\mathbf C_{\text D}$ is the $N_d\times N_d$ covariance matrix of the measurement errors $\bm e$, $\mathbf C_{\text{MM}}$ is the $N_m\times N_m$ autocovariance matrix of the input parameters in $\mathbf M$, $J_d^{\text{max}}$ and $J_m^{\text{max}}$ are the maximum values of $J_d(\bm d)$ and $J_m(\bm m)$, respectively. In earlier implementations of ILUES in~\citet{zhang2018}, the local ensemble of $\bm m_i$ is the $N_l=\alpha N_e,\;(\alpha\in(0,1])$ samples with the $N_l$ smallest $J$ values, i.e. $\mathbf M_i^l=[\bm m_{i,1},\ldots,\bm m_{i,N_l}]$, where $\alpha$ is the ratio between the sample sizes in $\mathbf M_i^l$ and $\mathbf M$. In this study, the local ensemble is determined using a roulette wheel selection operator~\citep{Lipowski2012} based on the $J$ values, in which the selection probability of the $i$-th individual is given as $P_i=\rho_i/\sum_{j=1}^{N_e}\rho_j$, $i=1,\ldots,N_e$, where $\rho_j=1/J(\bm m_j)$. This selection strategy may enhance the exploration of the parameter space in comparison to simply selecting the samples with the smallest $J$ values.
 
When assimilating the observation data, ILUES uses an inflated covariance matrix of the measurement errors to mitigate the large influence data mismatch at early iterations. The inflated covariance matrix is given by $\tilde{\mathbf C}_{\rm D}=\beta\mathbf C_{\rm D}$ and the factor $\beta$ is commonly chosen to be equal to the number of iterations $N_{\text{iter}}$~\citep{Emerick2012,Emerick2013,zhang2018}. Let $n$ denote the iteration index ($n=0,\ldots,N_{iter}$), we first update the local ensemble of each sample $\bm m_i^n$ ($i=1,\ldots,N_e$), i.e., $\mathbf M_i^{l,n}$, by using the usual ES update scheme~\citep{Emerick2013,zhang2018}:
 \begin{linenomath*}
 \begin{equation}\label{eq:es}
     \bm m_{i,j}^{n+1}=\bm m_{i,j}^{n}+\mathbf{C}_{\text{MD}}^{l,n}\big(\mathbf{C}_{\text{DD}}^{l,n}+\tilde{\mathbf C}_{\rm D}\big)^{-1}\big[\bm{d}_j-f(\bm{m}_{i,j}^n)\big],
 \end{equation}
 \end{linenomath*}
 for $j=1,\ldots,N_l$. Here $\mathbf{C}_{\text{MD}}^{l,n}$ is the cross-covariance matrix between $\mathbf M_i^{l,n}$ and $\mathbf D_i^{l,n}=\big[f(\bm m_{i,1}^n),\ldots,f(\bm m_{i,N_l}^n)\big]$, $\mathbf{C}_{\text{DD}}^{l,n}$ is the autocovariance matrix of $\mathbf D_i^{l,n}$, and $\bm d_j=\bm d+\tilde{\mathbf C}_{\rm D}^{1/2}\bm r_{N_d}$, $\bm r_{N_d}\sim \mathcal N(\bm 0,\mathbf I_{N_d})$, is the  $j$-th realization of the measurements. The update of $\bm m_i^n$ is then randomly selected from its updated local ensemble $\mathbf M_i^{l,{n+1}}=[\bm m_{i,1}^{n+1},\ldots,\bm m_{i,N_l}^{n+1}]$, say, $\bm m_{i,j}^{n+1}$ $(j=1,\ldots,N_l)$. To accelerate the convergence of the ILUES algorithm, we add an additional accept-reject step to accept the updated sample with an acceptance rate:
 \begin{linenomath*}
 \begin{equation}\label{eq:acc-rate}
     r_{\text{acc}}=\text{min}\Big\{1,\exp{\big[-\frac{1}{2}\big(J_d(\bm m_{i,j}^{n+1})-J_d(\bm m_i^n)\big)\big]}\Big\}.
 \end{equation}
 \end{linenomath*}
 Repeat this process for each sample $\bm m^n\in\mathbf M^n$ and the updated global ensemble  $\mathbf M^{n+1}=[\bm m_1^{n+1},\ldots,\bm m_{N_e}^{n+1}]$ is obtained. The ILUES algorithm is summarized in Algorithm~\ref{algor:ILUES}.

 \begin{algorithm}[htb]
\caption{Iterative local updating ensemble smoother. RWS:  roulette wheel selection.}
\label{algor:ILUES}
\begin{algorithmic}[1]

\Require Iteration number $N_{\rm iter}$, ensemble size $N_e$, local ensemble size factor $\alpha$.
\State $n\gets 0$. \Comment{Initialize the iteration index}
\State $N_l\gets\alpha N_e$.
\State Generate the initial input ensemble $\mathbf M^n=[\bm m_1^n,\ldots,\bm m_{N_e}^n]$ from the prior distribution.

\State Generate the initial output ensemble $\mathbf D^n=[f(\bm m_1^n),\ldots,f(\bm m_{N_e}^n)]$ by running the forward model.

\For{$n=1,\ldots,N_{\rm iter}$} \Comment Iterative data assimilation
  \For{$i=1,\ldots,N_e$} \Comment Update each sample using its local ensemble
     \State Given $\bm m_i^n$, compute the $J$ values for samples in $\mathbf M^n$ using equation~(\ref{eq:J-value}).
     
     \State Choose the local ensemble of $\bm m_i^n$, i.e., $\mathbf M_i^{l,n}=[\bm m_{i,1}^n,\ldots,\bm m_{i,N_l}^n]$, using RWS based on the $J$ values.
     
     \State Obtain the updated local ensemble $\mathbf M_i^{l,n+1}=[\bm m_{i,1}^{n+1},\ldots,\bm m_{i,N_l}^{n+1}]$ using equation~(\ref{eq:es}).
     
     \State Randomly draw a sample $\bm m_{i,j}^{n+1}\in \mathbf M_i^{l,n+1}$ and run the forward model $f(\bm m_{i,j}^{n+1})$.
     
     \State Compute $r_{\rm acc}$ in equation~(\ref{eq:acc-rate}) and generate a random number $z$ from $\mathcal U[0,1]$.
     
     \State  $\bm m_i^{n+1}=
    \begin{cases}
        \bm m_{i,j}^{n+1},& z\leq r_{acc}\\
      \bm m_i^n,& z> r_{acc}
    \end{cases}$ and $f(\bm m_i^{n+1})=
    \begin{cases}
        f(\bm m_{i,j}^{n+1}),& z\leq r_{acc}\\
      f(\bm m_i^n),& z> r_{acc}
    \end{cases}$.
     
  \EndFor
  \State \textbf{end for}
  \State $\mathbf M^{n+1}=[\bm m_1^{n+1},\ldots,\bm m_{N_e}^{n+1}]$. \Comment The updated input ensemble
  
  \State $\mathbf D^{n+1}=[f(\bm m_1^{n+1}),\ldots,f(\bm m_{N_e}^{n+1})]$.   \Comment The updated output ensemble
  
\EndFor
\State \textbf{end for}
\State \textbf{return} $\mathbf M^{N_{\rm iter}}$ and $\mathbf D^{N_{\rm iter}}$\Comment{The final input and output ensembles}
\end{algorithmic}
\end{algorithm}
 
 For high-dimensional inverse problems, large ensemble size and iteration number are required for ILUES to obtain converged and reliable inversion results~\citep{zhang2018}, leading to a large computational burden in forward model executions. To improve the computational efficiency, an accurate but fast-to-evaluate surrogate model constructed using a deep autoregressive neural network is used in this work to substitute the forward model in the ILUES algorithm.

\subsection{Deep Convolutional Neural Networks}
 DNNs are networks with a multi-layer structure~\citep{Goodfellow-et-al-2016}. Generally, given an input $\mathbf{x}$, the output of the $l$-th layer of a $L_{\rm{N}}$-layer network is given by
 \begin{linenomath*}
 \begin{equation}
     \mathbf{z}^{(l)}=\bm{\zeta}_l(\mathbf{z}^{(l-1)})=\eta_l(\mathbf{W}^{(l)}\mathbf{z}^{(l-1)}+\bm{b}^{(l)}),\:\forall l\in\{1,\ldots,L_{\rm N}\},
 \end{equation}
 \end{linenomath*}
 where $l$ denotes the layer index, $\eta(\cdot)$ is a nonlinear activation function, $\mathbf{W}^{(l)}$ and $\bm{b}^{(l)}$ are the weight matrix and bias vector, respectively, and $\mathbf{z}^0=\mathbf{x}$. In this work, the bias term is not considered, thus $\bm{\zeta}_l(\mathbf{z}^{(l-1)})=\eta_l(\mathbf{W}^{(l)}\mathbf{z}^{(l-1)})$.
 
 For DNNs, fully-connected networks may lead to an extremely large number of network parameters. The convolutional network architecture is commonly employed to greatly reduce the network parameters due to its sparse-connectivity and parameter-sharing properties. In addition,  convolutional neural networks are particularly suited for image-like data processing as the spatially-local dependence of the data is explicitly considered. More specifically, the layers of a convolutional neural network are organized into feature maps (matrices). Consider a two-dimensional input $\mathbf{x}\in\mathbb{R}^{H\times W}$, a convolutional layer, $\bm \zeta$, is obtained by employing a series of $q=1,\ldots,N_{\rm f}$ filters $\boldsymbol\omega^q \in\mathbb R^{k'_1\times k'_2}$, where $k'$ is referred to as kernel size, to evolve an input pixel $x_{u,v}$ to obtain the feature value $\zeta_{u,v}^q(x_{u,v})$ at location $(u,v)$ by
 \begin{linenomath*}
 \begin{equation}
    \zeta_{u,v}^q(x_{u,v})=f\left(\sum_{i=1}^{k'_1}\sum_{j=1}^{k'_2}\omega_{i,j}^q x_{u+i,v+j}\right).
 \end{equation}
 \end{linenomath*}
 This results in a convolutional layer consisting of $N_{\rm{f}}$ feature maps. The relationship between the output feature size $H_{\rm{fy}}\:(W_{\rm{fy}})$ and input feature size $H_{\rm{fx}}\:(W_{\rm{fx}})$ is given by~\citep{dumoulin2016}
 \begin{linenomath*}
 \begin{equation}\label{eq:featureSize}
     H_{\rm{fy}}=\left\lfloor\frac{H_{\rm{fx}}+2p-k'}{s}\right\rfloor+1,
 \end{equation}
 \end{linenomath*}
 where $s$ is the stride and $p$ is the zero padding that determine the distance between two successive moves of the filter and the padding of the borders of the input image with zeros for size preservation, respectively. Here,  $\left\lfloor\cdot\right\rfloor$ represents the floor function.
 
\subsection{Deep Autoregressive Neural Networks for Surrogate Modeling}
 
 In the DNN method considered here, an autoregressive strategy is proposed to improve the network's performance in predicting the    time-dependent response of the forward model with a time-varying source term. We adopt the dense convolutional encoder-decoder network architecture proposed in our previous work~\citep{mo2018,zhu2018}. It transforms the surrogate modeling task to an image-to-image regression problem to leverage the good properties of convolutional networks in image processing. In the following subsections, we briefly introduce the image-to-image regression strategy, the network architecture, and the deep autoregressive neural network. For the sake of completeness, in section~\ref{sec:image2image}, we summarize some elements of the work in~\citet{zhu2018} and~\citet{mo2018}. 

\subsubsection{Surrogate Modeling as Image-to-Image Regression}\label{sec:image2image}
 Assume that the governing equations for contaminant transport defined in section~\ref{sec:GovE} are solved over a two-dimensional rectangle domain of size $H\times W$, where $H$ and $W$ denote the grid resolution along the two axes of the domain. The model input-output relationship can then be described as a mapping of the form:
 \begin{linenomath*}
 \begin{equation}
    f:\mathbb R^{d_x\times H\times W}\to\mathbb R^{d_y\times H\times W},
 \end{equation}
 \end{linenomath*}
 where $d_x$ and $d_y$ denote the dimension of the inputs and outputs at one grid, respectively. It is straightforward to generalize to a three-dimensional spatial domain by adding an extra depth axis, i.e., $\textbf x\in\mathbb R^{d_x\times D\times H\times W}$ and  $\textbf y\in\mathbb R^{d_y\times D\times H\times W}$.  The surrogate modeling task to approximate such an input-output mapping can be transformed to an image-to-image regression problem. To this end, the inputs and outputs are organized as image-like data, e.g., $\mathbf{x}\in\mathbb{R}^{d_x\times H\times W}$ represents $d_x$ images with a resolution of $H\times W$. The image regression problem can be efficiently solved using a deep dense convolutional encoder-decoder network architecture~\citep{mo2018,zhu2018}.  
 
 In the convolutional encoder-decoder network, the regression between high-dimensional images is performed through a coarse-refine process, i.e., an encoder is used to extract high-level coarse features from the input images, which are then refined to output images through a decoder. Additionally, to alleviate the training data-intensive issue of DNNs, the network is fully-convolutional~\citep{long2015} without containing any fully-connected layers and a densely connected convolutional network structure called `dense block'~\citep{huang2017} is employed. The dense block introduces connections between non-adjacent layers to enhance the information propagation through the network. Consequently, the $l$-th layer receives the feature maps of all preceding layers in the dense block i.e. $\mathbf{z}^{(l)}=\bm{\zeta}_l([\mathbf{z}^{(l-1)},\ldots,\textbf z^0])$. A dense block contains two design parameters, the number of internal layers $L$ and the number of output features $R$ of each layer (also referred to as growth rate). Each layer contains three consecutive operations: batch normalization (BN)~\citep{ioffe2015batch}, followed by rectified linear unit (ReLU)~\citep{Goodfellow-et-al-2016} and convolution (Conv)~\citep{Goodfellow-et-al-2016}.
 
\begin{figure}[h!]
 \centering
 \includegraphics[width=0.55\textwidth]{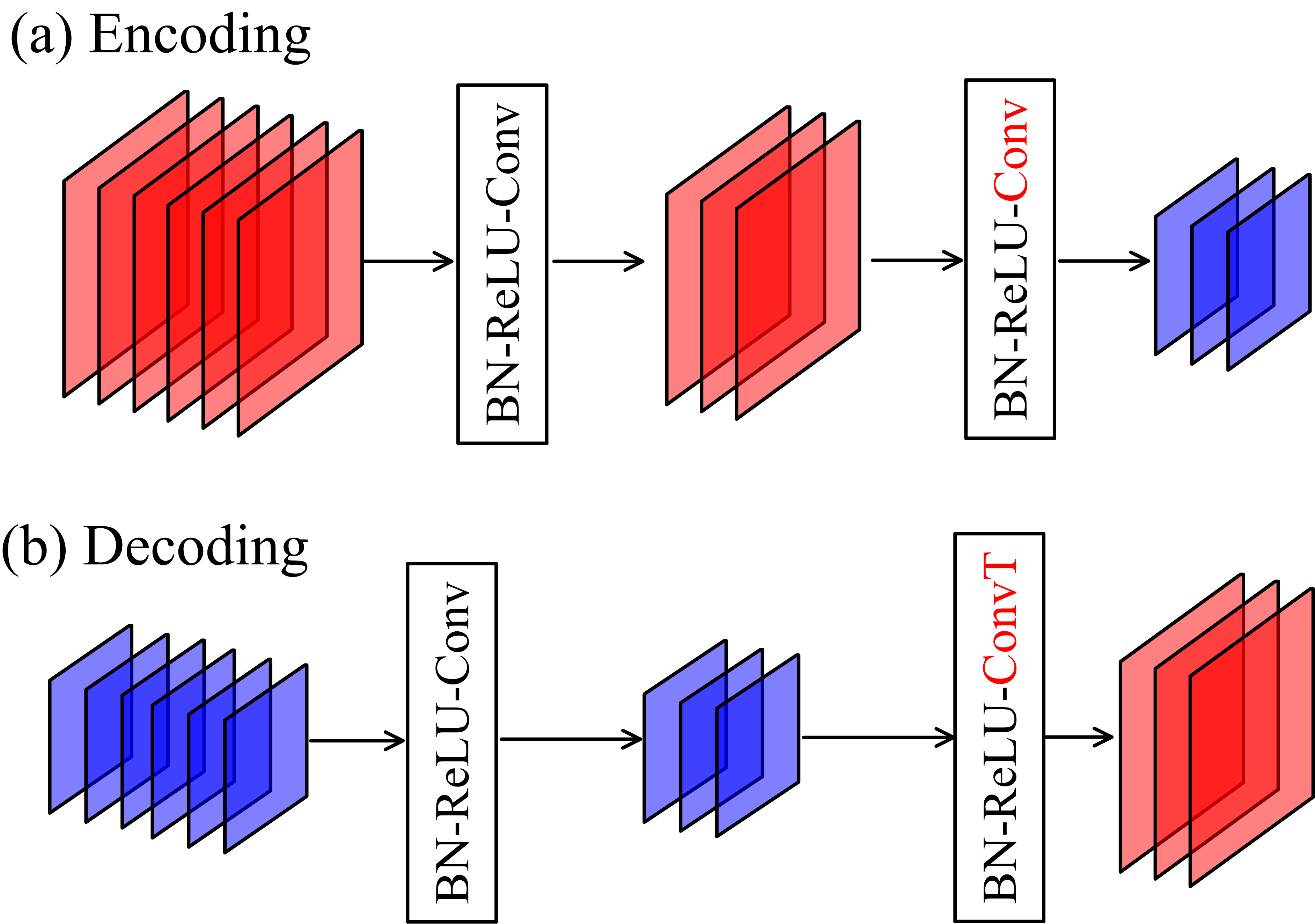}
    \caption{(a) An encoding layer and (b) a decoding layer, both of which contain two convolutions (Convs). The first Conv in both (a) and (b) halves the number of feature maps while keeps their size the same; the second Conv in (a) halves the feature size (downsampling) and the transposed Conv (ConvT) in (b) doubles the feature size (upsampling).}
 \label{fig:transition-layer}
 \end{figure}

 Transition layers, which are referred to as encoding layers in the encoder and as decoding layers in the decoder, are placed between  two adjacent blocks to change the feature size via (transposed) convolution~\citep{dumoulin2016} during the coarse-refine process and to avoid feature maps explosion (due to the concatenation of feature maps in dense blocks). They halve the size of feature maps in the encoder (i.e., downsampling), while double the size in the decoder (i.e., upsampling). Both the encoding and decoding layers halve the number of feature maps. An illustration of the encoding and decoding layers is given in Figure~\ref{fig:transition-layer}. 
 
\begin{figure}[h!]
 \centering
 \includegraphics[width=0.95\textwidth]{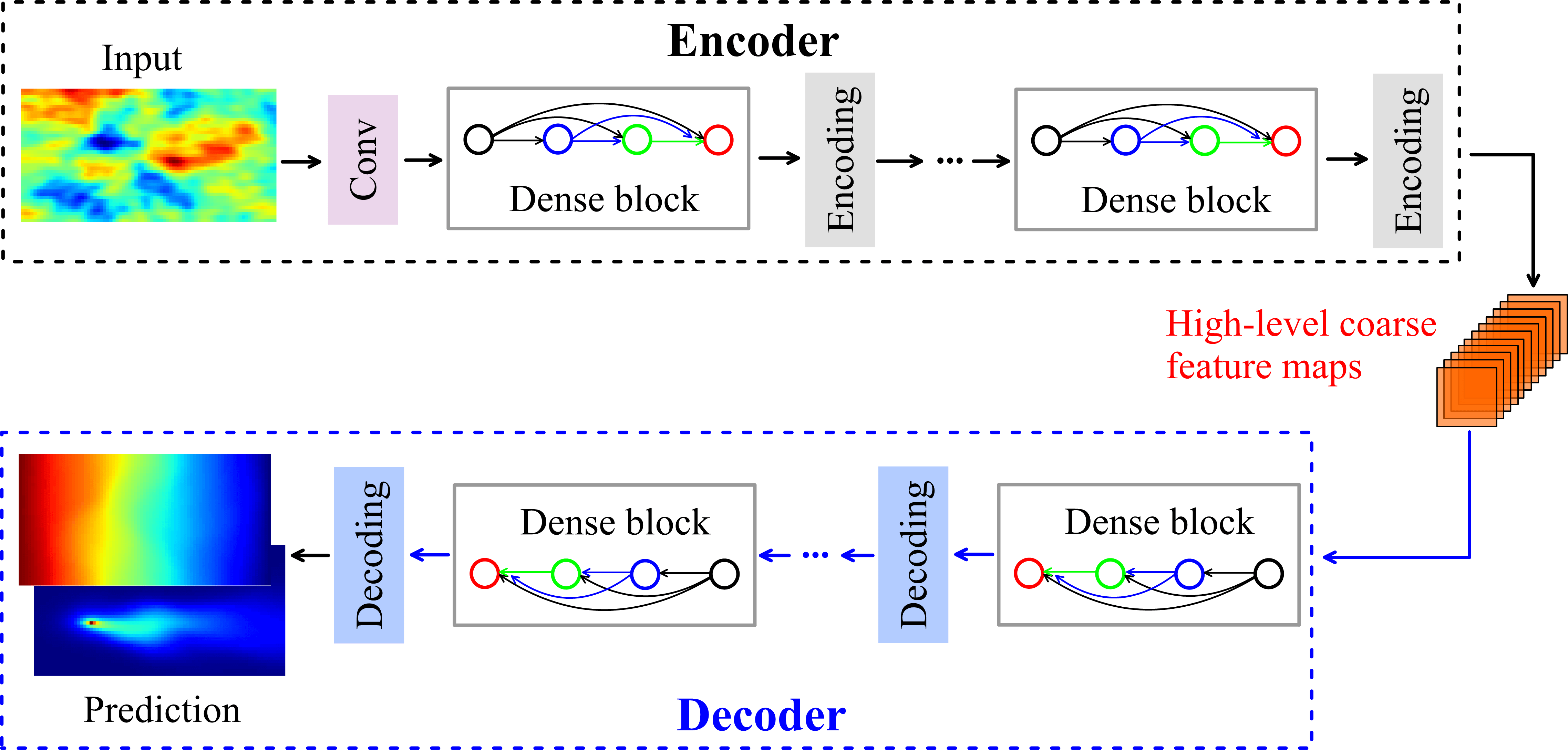}
     \caption{Illustration of the deep dense convolutional encoder-decoder network architecture. The encoder operates on the input image to extract a set of high-level coarse feature maps (red boxes), which are then fed into the decoder to eventually reconstruct the output images.}
 \label{fig:DCEDN}
 \end{figure}

 The deep dense convolutional encoder-decoder network architecture is depicted in Figure~\ref{fig:DCEDN}. It is an alternating cascade of dense blocks and transition layers. In the encoding path, a convolution layer is first used to extract feature maps from the raw input images. The extracted feature maps are then passed through an alternating cascade of dense blocks and encoding layers. The last encoding layer outputs high-level coarse feature maps, as shown with red boxes, which are subsequently fed into the decoder. The decoder is composed of an alternation of dense blocks and decoding layers, with the last decoding layer reconstructing the output images. The network can easily handle multiple input and output images without any modifications of the network architecture~\citep{mo2018,zhu2018}.
 
 \subsubsection{Input Image for the Source Term} \label{sec:image4source}
 As introduced in section~\ref{sec:image2image}, the inputs are treated as images (i.e., matrices) in the network. For the vectorial input source location $\bm{S}_l=(S_{lx},S_{ly})$ and time-varying source strength $\bm{S}_s=\{S_{sj}\}_{j=1}^{n_t}$, one can simply treat each parameter as an image with a constant value equaling to the parameter value at all pixels, e.g., the input image for $S_{lx}$ is $\mathbf S_{lx}\in\mathbb R^{H\times W}$ with $\mathbf S_{lx,(u,v)}=S_{lx}$ for $u=1,\ldots,H$, $v=1,\ldots,W$. Nevertheless, as the spatial dependence of the data is rigorously considered in convolutional neural networks, the input image for the source term (including the location and strength) at the $j$-th time segment, $\mathbf S_j\in\mathbb R^{H\times W}$, is organized as
 \begin{linenomath*}
 \begin{equation}
    S_{j,(u,v)}=
    \begin{cases}
        S_{sj},& u=u_s,v=v_s\\
        0,& u\neq u_s,v\neq v_s
    \end{cases},
 \end{equation}
 \end{linenomath*}
 for $u=1,\ldots,H$, $v=1,\ldots,W$. Here $S_{j,(u,v)}$ is the value of matrix $\mathbf S_j$ at index $(u,v)$, and $(u_s,v_s)$ is the index corresponding to the source location $\bm{S}_l=(S_{lx},S_{ly})$ in the image. The resulting image thus clearly reflects the spatial information on the source term, which may be beneficial to the network's predictive accuracy.

\subsubsection{Deep Autoregressive Neural Networks for Time-Varying Processes}
\begin{figure}[h!]
 \centering
    \includegraphics[width=.95\textwidth]{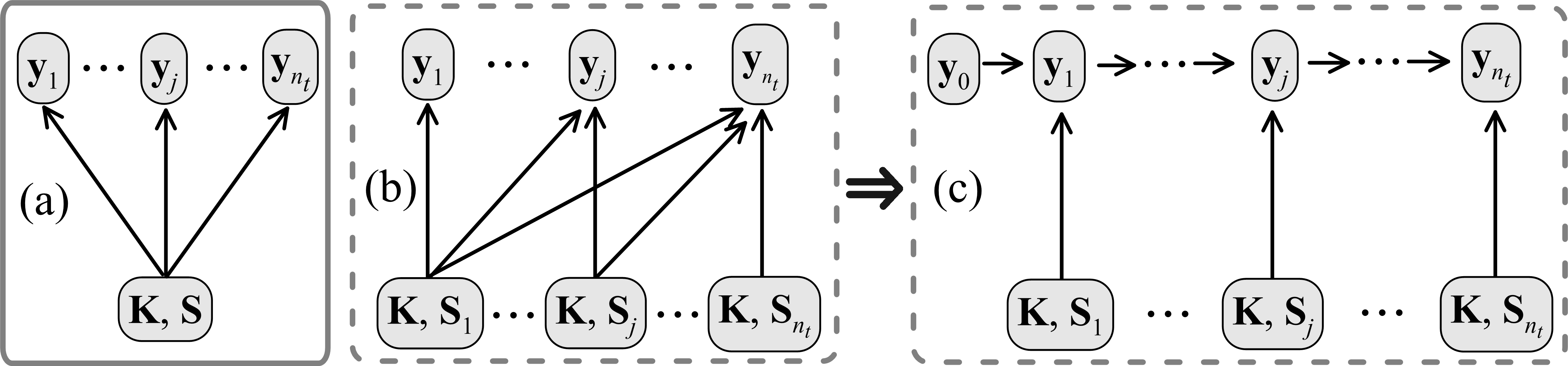}
    \caption{(a)-(b): Relationship between the uncertain model inputs (the conductivity field $\mathbf{K}$ and source term $\mathbf S$) and the time serials output concentration fields, $\mathbf{y}_j,\,j=1,\ldots,\,n_t$. The source term is constant in (a) and time-varying in (b). In (b), $\mathbf{y}_j$ depends on $\mathbf{K}$ and $\mathbf S_k$, $k=1,\ldots,\,j$. (c): The model input-output relationship  is represented using an autoregressive model in which the output at the previous time step $\mathbf y_{j-1}$, $j=1,\ldots,n_t$, is treated as input to predict the current output $\mathbf y_j$, where $\mathbf{y}_0$ is the known initial state.}
    \label{fig:data_reorg}
 \end{figure}

 Without loss of generality, here we use $\mathbf y\in\mathbb{R}^{n_t\times H\times W}$ to denote the output concentration fields at $n_t$ time steps. Figure~\ref{fig:data_reorg} depicts the dependence of the output concentration field at the $j$-th time step $\mathbf{y}_j\in\mathbb{R}^{H\times W}$, $j=1,\ldots,n_t$, on the model input conductivity field $\mathbf{K}$ and source term $\mathbf{S}$. When the source strength is constant, the model input-output relationship is simple as the outputs at different time steps $\mathbf{y}_j\in\mathbb{R}^{H\times W},\,j=1,\ldots,\,n_t$, depend on the same inputs $\mathbf{K}$ and $\mathbf S$ (Figure~\ref{fig:data_reorg}a). It is therefore straightforward to construct a network for such a system by simply considering two input channels for $\mathbf{K}$ and $\mathbf{S}$, respectively, and $n_t$ output channels for $\mathbf y\in\mathbb{R}^{n_t\times H\times W}$. However, when the source strength is time-varying, i.e., $\mathbf S=[\mathbf S_1,\ldots,\mathbf S_{n_t}]$, the model input-output relationship is complicated. As shown in Figure~\ref{fig:data_reorg}b, apart from the time-independent input conductivity field $\mathbf K$, the output concentration fields at different time steps depend on different source terms in $\{\mathbf S_j\}_{j=1}^{n_t}$. For example, $\mathbf y_1$ depends only on one source term $\mathbf S_1$, whereas $\mathbf{y}_j$, depends on $\mathbf{S}_k$, $k=1,\ldots,\,j$. That is, the output concentration field at an arbitrary time step is only affected by the previously released contaminant. In this case, it may no longer be effective to construct the network in the same way as in the constant-strength case, i.e., considering $n_t+1$ input channels (one for $\mathbf{K}$ and the other $n_t$ channels for $\mathbf S=[\mathbf S_1,\ldots,\mathbf S_{n_t}]$) and $n_t$ output channels for $\mathbf y\in\mathbb{R}^{n_t\times H\times W}$. In this case, the relationship between the input source term at different time segments and the outputs is not clearly captures by the network architecture. 
 
 We propose to represent the time-varying process using an autoregressive model, as follows:
 \begin{linenomath*}
 \begin{equation}
 \label{eq:autoreg}
     \mathbf{y}_j=f(\mathbf{K},\mathbf{S}_j,\mathbf{y}_{j-1}).
 \end{equation}
 \end{linenomath*}
 In this model, the current output $\mathbf{y}_j$, $j=1,\ldots,n_t$, depends only on the previous output $\mathbf{y}_{j-1}$, the input source term at the last time interval $\mathbf{S}_j$, and the time-independent input conductivity field $\mathbf K$, where $\mathbf{y}_0$ is the known initial state. The information about the  past inputs and states is contained in $\mathbf{y}_{j-1}$. Therefore, in the autoregressive network, we reorganize the data obtained from one forward model execution, i.e. $(\mathbf{K},\mathbf S_1,\ldots,\mathbf S_{n_t};\mathbf y_1,\ldots,\mathbf y_{n_t})$, as $n_t$ training samples:
 \begin{linenomath*}
 \begin{equation}
     (\mathbf{K},\mathbf S_1,\ldots,\mathbf S_{n_t};\mathbf y_1,\ldots,\mathbf y_{n_t})\Rightarrow \big\{(\mathbf{K},\mathbf S_j,\mathbf{y}_{j-1};\,\mathbf{y}_j)\big \}_{j=1}^{n_t}.
 \end{equation}
 \end{linenomath*}
 In this way, the relationship between the time-dependent model inputs and outputs is clearly captured by the network as shown in Figure~\ref{fig:data_reorg}c. Consequently, given the results of $N$ forward model runs, $\big\{(\mathbf{x}^i,\,\mathbf{y}^i)\big\}_{i=1}^{N} = \big\{(\mathbf{K}^i,\mathbf{S}^i;\,\mathbf{y}^i)\big\}_{i=1}^{N}$, where $\mathbf S^i=\{\mathbf S_1^i,\ldots,\mathbf S_{n_t}^i\}$ and $\mathbf y^i=\{\mathbf y_1^i,\ldots,\mathbf y_{n_t}^i\}$, the training data are organized as:
 \begin{linenomath*}
 \begin{equation}
     \label{eq:data-reorg}
     \big\{(\mathbf{K}^i,\mathbf{S}^i;\,\mathbf{y}^i)\big\}_{i=1}^{N}\Longrightarrow \big\{(\mathbf{K}^i,\mathbf{S}_j^i,\mathbf{y}_{j-1}^i;\,\mathbf{y}_j^i)\big \}_{i=1,j=1}^{N,n_t}= \big\{(\tilde{\mathbf{x}}^m,\,\mathbf{y}^m)\big\}_{m=1}^{Nn_t},
 \end{equation}
 \end{linenomath*}
 where $\mathbf{y}_j^i\in\mathbb R^{H\times W}$ denotes the $i$-th model run's simulation output at the $j$-th time step, $\tilde{\mathbf{x}}^m=(\mathbf{K}^i,\mathbf{S}_j^i,\mathbf{y}_{j-1}^i)$, $\mathbf{y}^m=\mathbf{y}_j^i$, and $m$ is a re-index for $(i,j)$. 

\begin{figure}[h!]
 \centering
    \includegraphics[width=.5\textwidth]{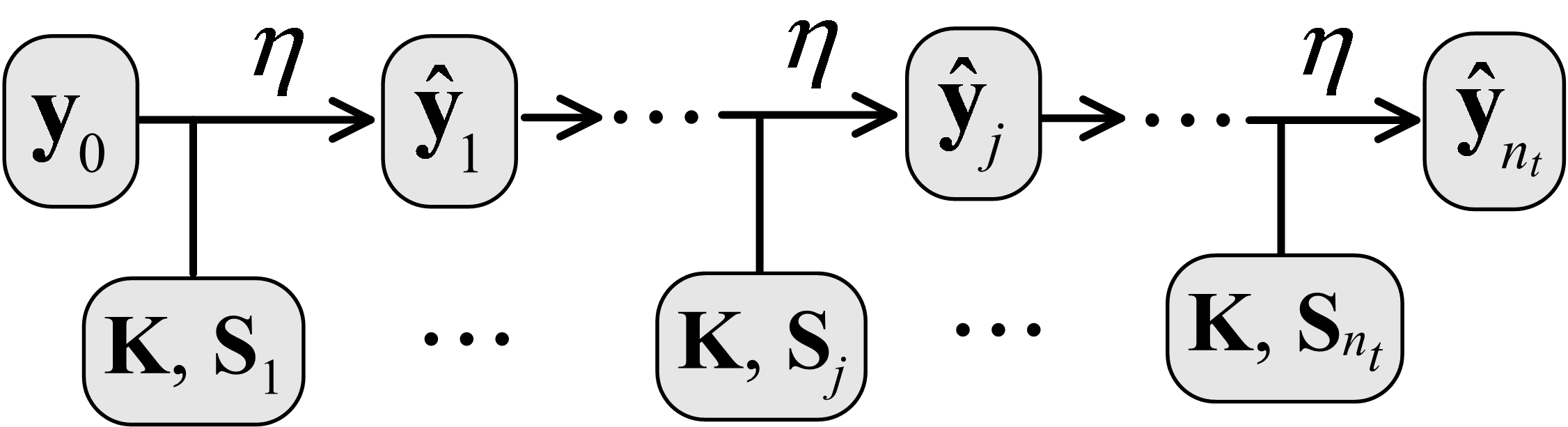}
    \caption{Sequential prediction of the model output $\mathbf{y}$ from time $t_1$ to time $t_{n_t}$ given an arbitrary input conductivity $\mathbf{K}$ and time-varying source term $\mathbf{S}=\{\mathbf S_j\}_{j=1}^{n_t}$ using the trained autoregressive neural network $\eta$. $\mathbf{y}_0$ is the known initial state.}
    \label{fig:surrogate_pred}
 \end{figure}

 Once the autoregressive neural network $\eta$ is trained using the training data $(\tilde{\mathbf{X}},\mathbf{Y})=\big\{(\tilde{\mathbf{x}}^m,\mathbf{y}^m)\big\}_{m=1}^{Nn_t}$, the predicted output fields $\hat{\mathbf{y}}$ for an arbitrary input conductivity $\mathbf{K}$ and source term $\mathbf{S}$ can be obtained by following the procedure illustrated in Figure~\ref{fig:surrogate_pred}. Each output field $\mathbf y_j$ at the $n_t$ time steps is sequentially predicted using its previous state $\mathbf y_{j-1}$ combined with $\mathbf{K}$ and $\mathbf S_j$ as the input. The network $\eta$ first predicts the output field $\hat{\mathbf{y}}_1$ using the input $\tilde{\mathbf{x}}=(\mathbf{K},\mathbf{S}_1,\mathbf{y}_0)$; as such, $\hat{\mathbf{y}}_2$ is obtained by using the input $\tilde{\mathbf{x}}=(\mathbf{K},\mathbf{S}_2,\hat{\mathbf{y}}_1)$. This procedure is repeated  $n_t$ times and the predicted output fields $\hat{\mathbf{y}}=\{\hat{\mathbf{y}}_j\}_{j=1}^{n_t}$ for the input $\mathbf{x}=(\mathbf{K},\mathbf{S})$ are obtained. Note that the above procedure is computationally cheap as it does not involve any forward model evaluations. The effectiveness of the autoregressive strategy in improving the network approximation accuracy will be illustrated in section~\ref{sec:approx-accuracy}.
 
\subsubsection{Network Training}
 During network training, we derive $\hat{\mathbf{y}}=\eta(\mathbf x,\boldsymbol\theta)$ to match $\mathbf{y}=f(\mathbf x)$ by learning the values of all the network trainable parameters $\boldsymbol\theta$. Given a set of training samples $\{(\tilde{\mathbf{x}}^m,\mathbf{y}^m)\}_{m=1}^{Nn_t}$, the network is trained to minimize a certain loss. The regularized $L_1$-norm loss function is used, as represented by
 \begin{linenomath*}
 \begin{equation}
    J(\boldsymbol\theta)=\mathcal{L}(\hat{\mathbf{y}},\mathbf{y},\boldsymbol\theta)=\frac{1}{Nn_t}\sum_{m=1}^{Nn_t}|\eta(\tilde{\mathbf{x}}^m,\boldsymbol\theta)-\mathbf{y}^m|+\frac{\omega_d}{2}\boldsymbol\theta^\top\boldsymbol\theta,
 \label{eq:loss}
 \end{equation}
 \end{linenomath*}
 where  $\omega_d$ is a regularization coefficient, also called weight decay.

 Stochastic gradient descent (SGD) is used as the optimizer for parameter learning~\citep{Goodfellow-et-al-2016}. Gradient descent requires computing
 \begin{linenomath*}
 \begin{equation}
    -\nabla_{\boldsymbol\theta}J(\boldsymbol\theta)=\frac{1}{Nn_t}\sum_{m=1}^{Nn_t}-\nabla_{\boldsymbol\theta}\mathcal{L}(\hat{\mathbf{y}}^m,\mathbf{y}^m,\boldsymbol\theta).
 \end{equation}
 \end{linenomath*}
 The computational cost of this operation may be very high when $Nn_t$ is large. In SGD, a minibatch strategy is used to reduce the computational burden, in which only a minibatch of samples  ($M< Nn_t$)  randomly drawn from the training set is used~\citep{Goodfellow-et-al-2016}. Then the estimate of the gradient is given as
 \begin{linenomath*}
 \begin{equation}
    \bm{g}=\frac{1}{M}\sum_{m=1}^{M}-\nabla_{\boldsymbol\theta}\mathcal{L}(\hat{\mathbf{y}}^m,\mathbf{y}^m,\boldsymbol\theta).
 \end{equation}
 \end{linenomath*}
 The SGD then follows the estimated gradient downhill:
 \begin{linenomath*}
 \begin{equation}
    \boldsymbol\theta\leftarrow\boldsymbol\theta+\varepsilon\bm{g},
 \end{equation}
 \end{linenomath*}
 where $\varepsilon$ is the learning rate. Various SGD algorithms are available. In our network, the widely used Adam algorithm~\citep{kingma2014} is used.

\section{Application}\label{sec:application}

\subsection{Contaminant Transport Model}
\begin{figure}[h!]
 \centering
    \includegraphics[width=.6\textwidth]{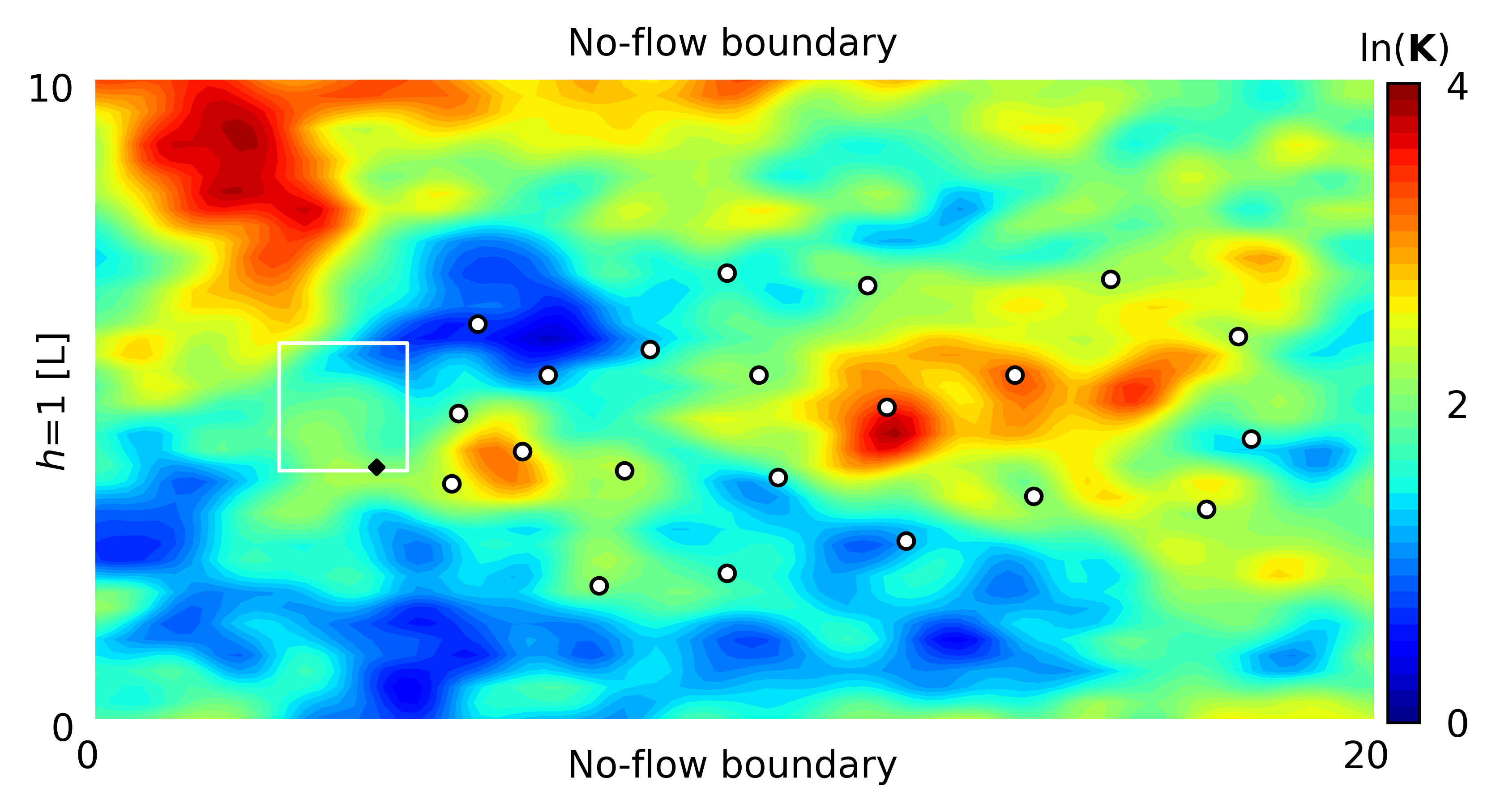}
    \caption{Flow domain of the contaminant transport model and the reference log-conductivity field considered in the inverse problem. The black diamond denote the reference contaminant source location and the white rectangle represents the potential area of the source. The $21$ black circles denote the measurement locations.}
    \label{fig:conceptual_model}
 \end{figure}

 In this section, a synthetic two-dimensional contaminant transport model revised after that of~\citet{zhang2015} is used to illustrate the performance of the proposed method. As shown in Figure~\ref{fig:conceptual_model}, the spatial domain has a size of $10\;[\text{L}]\times20\;[\text{L}]$ with constant heads of $h=1\;[\text{L}]$ and $h=0\;[\text{L}]$ at the left and right boundaries, respectively; whereas the two lateral boundaries are assumed to be no-flow boundaries. The domain is discretized uniformly into $41\times81$ cells. During the time interval between $0\;[\text{T}]$ and $10\;[\text{T}]$, the contaminant is released with a time-varying strength from an unknown location within a potential area denoted by a white rectangle in Figure~\ref{fig:conceptual_model}. The source strength is taken as the mass-loading rate $[\text{MT}^{-1}]$, i.e., the mass released in unit time. The unknown source is characterized by seven parameters, including the location $\bm{S}_l=(S_{lx},S_{ly})$, time-varying strengths $\bm S_s=\{S_{sj}\}_{j=1}^5$ in five time segments $[t_{j-1}:t_j]$, where $t_j=2j\;[\text{T}],\;j=1,\ldots,5$. 
 
 In this case, we assume that the porosity, longitudinal and transverse dispersivities are known with constant values of $\theta=0.25$, $\alpha_L=1.0\;[\text{L}^2\text{T}^{-1}]$ and $\alpha_T=0.1\;[\text{L}^2\text{T}^{-1}]$, respectively. The hydraulic conductivity field is heterogeneous and assumed to be a log-Gaussian random field, i.e.,
 \begin{linenomath*}
 \begin{equation}
     K(\bm{s})=\exp{\big(G(\bm{s})\big)},\:G(\bm{s})\sim N \big( m(\bm{s}),C(\cdot,\cdot) \big),
 \end{equation}
 \end{linenomath*}
 with a $L_2$ norm exponential covariance function:
 \begin{linenomath*}
 \begin{equation}
     C(\bm{s},\bm{s}')=\sigma_G^2\exp{\left(-\sqrt{\left(\frac{s_x-s'_x}{\lambda_x}\right)^2+\left(\frac{s_y-s'_y}{\lambda_y}\right)^2}\right)},
     \label{eq:cov}
 \end{equation}
 \end{linenomath*}
 where $\bm{s}=(s_x,s_y)$ and $\bm{s}'=(s'_x,s'_y)$ are two arbitrary spatial locations, $m(\bm{s})=2$ is the constant mean, $\sigma_G^2=0.5$ is the variance, and $\lambda_x$ and $\lambda_y$ are the correlation lengths along the $x-$ and $y-$axes, respectively. In this example, we consider heterogeneous conductivity fields with a length scale of $\lambda/L_{\rm d}=0.2$ in both directions, where $L_{\rm d}$ is the domain size. Although the employed network architecture can effectively handle high-dimensional input fields even without using any dimensionality reduction techniques~\citep{mo2018,zhu2018}, an inverse problem with a high-input dimensionality will need a large number of forward model runs to obtain converged inversion results. Thus, to enable the implementation of the inversion method ILUES to relatively quickly obtain a converged solution, the Karhunen-Lo{\` e}ve expansion (KLE)~\citep{zhang2004} is employed to parameterize the log-conductivity field, as represented by
 \begin{linenomath*}
 \begin{equation}
     G(\bm{s})=m(\bm{s})+\sum_{i=1}^{\infty}\xi_i\sqrt{\lambda_i}\phi_i(\bm{s}),
 \end{equation}
 \end{linenomath*}
 where $\lambda_i$ and $\phi_i(\bm{s})$ are eigenvalues and eigenvectors of the convariance function, respectively, and $\xi_i\sim \mathcal N(0,1)$ are the KLE coefficients. The number of KLE terms required to preserve a specific percentage of the total field variance  is determined by the length scale. A random field with a small length scale will vary  highly over the domain and therefore a  large number of KLE terms is required to accurately represent this high-variability (i.e., high-heterogeneity) feature. For the highly-heterogeneous field considered, we keep the first $N_{\text{KL}}=679$ dominant KLE terms to preserve $95\%$ of the total field variance (i.e., $\sum_{i=1}^{N_{\text{KL}}}\lambda_i\slash\sum_{i=1}^{\infty}\lambda_i\approx0.95)$. As a result, in our inverse problem there are $N_m=686$ uncertain parameters, including $2$ source location parameters $(S_{lx},S_{ly})$, $5$ source strength parameters $\left\{S_{sj}\right\}_{j=1}^5$, and $679$ KLE coefficients $\left\{\xi_i\right\}_{i=1}^{679}$. 
 
 Although the dimensionality of the conductivity field $\mathbf{K}$ has been reduced from $3321$ (i.e., the number of spatial grids) to $679$ (i.e., the number of preserved KLE coefficients), in the ILUES algorithm, the input fed to the network is still the image-like conductivity field reconstructed from KLE. Figure~\ref{fig:surrogate_pred} shows the procedure for making predictions for an arbitrary realization of the uncertain parameters $\bm m=(\boldsymbol{\xi},\bm{S}_l,\bm{S}_s)$. We first generate the conductivity field $\mathbf{K}$ using KLE based on the given coefficients $\boldsymbol{\xi}$ and the input images $\mathbf S\in\mathbb R^{5\times H\times W}$ for the source term $(\bm{S}_l,\bm{S}_s)$ as introduced in section~\ref{sec:image4source}; then based on the generated $\mathbf{K}$ and $\mathbf S$ the predicted output fields $\hat{\mathbf{y}}$ are obtained using the trained deep neural network.
 
 \begin{figure}[h!]
 \centering
    \includegraphics[width=.6\textwidth]{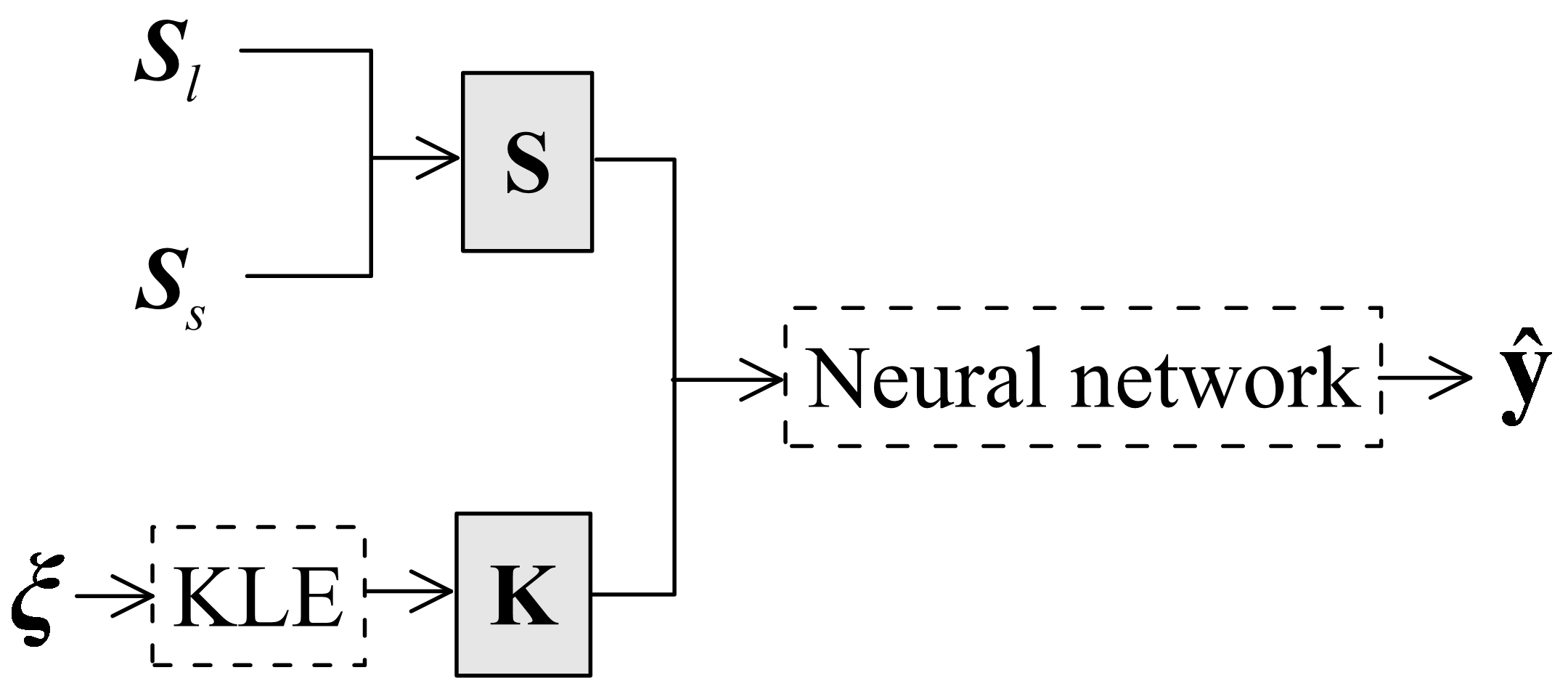}
    \caption{Illustration of the neural network prediction for an arbitrary uncertain input vector $\bm m=(\boldsymbol{\xi},\bm{S}_l,\bm{S}_s)$. The conductivity field $\mathbf{K}$ is first generated by KLE given the coefficient vector $\boldsymbol{\xi}$. This together with the input images $\mathbf S$ for the source term $(\bm{S}_l,\bm{S}_s)$ (see section~\ref{sec:image4source}) are used to predict the output fields $\hat{\mathbf{y}}$ using the trained neural network.}
    \label{fig:refK}
 \end{figure}
 
 The prior distributions of the $686$ unknown parameters are shown in Table~\ref{tab:prior}. The $7$ uncertain source parameters are assumed to be uniformly distributed within the given ranges listed in Table~\ref{tab:prior}. The $679$ KLE coefficients have a standard normal distribution, i.e., $\xi_i\sim\mathcal N(0,1)$. The randomly selected reference values for the $7$ source parameters and reference log-conductivity field are shown in Table~\ref{tab:prior} and Figure~\ref{fig:conceptual_model}, respectively. To infer the unknown parameters, the concentrations at $t=[2,4,6,8,10,12,14]$~[T] and hydraulic heads are collected at $21$ measurement locations denoted by the black circles in Figure~\ref{fig:conceptual_model}. Thus $N_d=21\times 8=168$ observations are available. The synthetic observations are generated by adding $5\%$ independent Gaussian random noise to the data generated by the forward model using the reference parameter values.

 \begin{table}[h!]
    \caption{Prior distributions and reference values of the $686$ uncertain parameters for the case study. The reference values for the KLE coefficients $\{\xi_i\}_{i=1}^{679}$ are not listed; the corresponding reference log-conductivity field is shown in Figure~\ref{fig:conceptual_model}.}
    \centering
    \label{tab:prior}
    \begin{tabular}{lcccccccc}
        \hline
        Parameter  & $S_{lx}$    & $S_{ly}$    & $S_{s1}$    & $S_{s2}$    & $S_{s3}$    & $S_{s4}$    & $S_{s5}$    & $\{\xi_i\}_{i=1}^{679}$ \\
        \hline
         Prior      & $\mathcal U[3,5]$ & $\mathcal U[4,6]$ & $\mathcal U[0,8]$ & $\mathcal U[0,8]$ & $\mathcal U[0,8]$ & $\mathcal U[0,8]$ & $\mathcal U[0,8]$ & $\mathcal N(0,1)$             \\
        Reference value & $4.5234$  & $4.0618$  & $6.5989$  & $1.0502$  & $1.8535$  & $6.5638$  & $2.9540$  & -  \\
        \hline
    \end{tabular}
 \end{table}

 \subsection{A Weighted Loss to Improve the Approximation of the Concentration Field}\label{sec:weighted-loss}
\begin{figure}[h!]
 \centering
    \includegraphics[width=\textwidth]{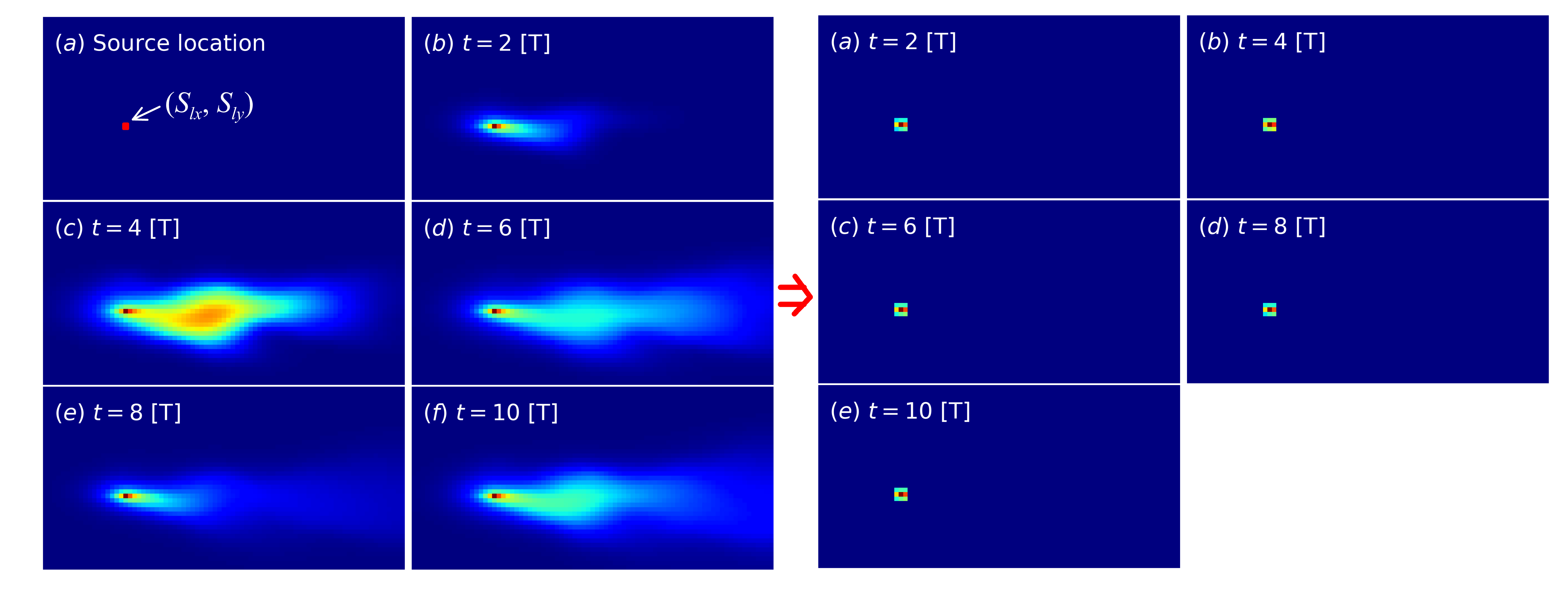}
    \caption{Left: The contaminant source release location (a) and the output concentration fields at $t=2,4,\ldots,10$ [T]. Right: The extracted concentrations at the source pixel and its surrounding eight pixels in the concentration fields. These concentration values are assigned an additional weight in the loss when training the network.}
    \label{fig:conc-cmax}
 \end{figure}

 The concentration field usually has a large gradient in the region near the source release location due to small dispersivity. This strong nonlinearity is difficult to model and often leads to large approximation errors~\citep{liao2013}. To improve the network's approximation accuracy in this region, in the loss function an additional weight $w_c$ is assigned to the source pixel (grid) and its surrounding eight pixels in the five concentration images at time $t_1$ to $t_5$ when the contaminant is released. As shown in Figure~\ref{fig:conc-cmax}, according to the source location $(S_{lx},S_{ly})$, the concentrations at the source pixel and its surrounding eight pixels in the concentration field are extracted. Let $\bm c_s$ denote the vector of the nine concentrations extracted from one concentration field, the weighted loss can then be written as
 \begin{linenomath*}
 \begin{equation}
    J'(\boldsymbol\theta)=J(\boldsymbol\theta)+J_c(\boldsymbol\theta)=J(\boldsymbol\theta)+w_c\sum_{i=1}^{Nn'_t}|\hat{\bm{c}}_{si}-\bm{c}_{si}|, 
    \label{eq:weighted-loss}
 \end{equation}
 \end{linenomath*}
 where $J(\boldsymbol\theta)$ is the $L_1$ loss as defined in equation~(\ref{eq:loss}), $n'_t=5$ is the number of time segments that the contaminant is released, and $\hat{\bm c}_s$ is the network's prediction for $\bm c_s$. A discussion on the choice of the value of weight $w_c$ is given in section~\ref{sec:approx-accuracy}.
 
 \subsection{Networks and Datasets}
 To assess the effectiveness of the autoregressive strategy in improving the network's approximation accuracy for the time series outputs of the contaminant transport model with a time-varying source term, and of the weighted loss in improving the characterization of the large-gradient concentration feature around the source release location, three types of networks are trained and compared:
 
 \texttt{Net}: Network without autoregression. No autoregression is used in this network. It is  trained using the original training samples without reorganizing the data using equation~(\ref{eq:data-reorg}). That is, one model execution results in one training sample. The $L_1$ loss function defined in equation~(\ref{eq:loss}) is used. 
 
 \texttt{AR-Net}: Autoregressive network. In this network, the time-varying process is represented using an autoregressive model as defined in equation~(\ref{eq:autoreg}) and the network is trained using the reorganized training data as defined in equation~(\ref{eq:data-reorg}). That is, one model execution results in $n_t$ reorganized training samples. The $L_1$ loss function defined in equation~(\ref{eq:loss}) is used.
 
 \texttt{AR-Net-WL}: Autoregressive network with the weighted loss. \texttt{AR-Net-WL} is similar to \texttt{AR-Net} except that it is trained using a weighted loss function as defined in equation~(\ref{eq:weighted-loss}). 
 
 To illustrate the convergence of the approximation error with respect to the training sample size, three training sample sets obtained from $N=400$, $1000$, and $1500$ model evaluations are generated. The concentration at $t=[2,4,6,8,10,12,14]$~[T] and the hydraulic heads are taken as the observations for the inverse problem. Thus we collect the output concentration fields at the $n_t=7$ time instances $\{\mathbf c_j\}_{j=1}^{n_t}$ and the output hydraulic field $\mathbf h$ to train the network. The input to network is the hydraulic conductivity field $\mathbf K$ and the images $\{\mathbf S_j\}_{j=1}^{n_t}$ for the time-varying source term (see section~\ref{sec:image4source}). To assess the quality of the trained networks, $N_{\rm{test}}=500$ model runs are performed to generate the test samples. The quality is evaluated using the coefficient of determination ($R^2$) and the root mean squared error (RMSE) which are defined as
 \begin{linenomath*}
 \begin{equation}
     R^2=1-\frac{\sum_{i=1}^{N_{\rm{test}}}||\textbf y^i-\hat{\textbf y}^i||_2^2}{\sum_{i=1}^{N_{\rm{test}}}||\textbf y^i-\bar{\textbf y}||_2^2},
 \end{equation}
 \end{linenomath*}
 and
 \begin{linenomath*}
 \begin{equation}
     \text{RMSE}=\sqrt{\frac{1}{N_{\rm{test}}}\sum_{i=1}^{N_{\rm{test}}}||\mathbf{y}^i-\hat{\mathbf{y}}^i||_2^2},
 \end{equation}
 \end{linenomath*}
 respectively. Here $\mathbf{y}^i,\:\hat{\mathbf{y}}^i\in \mathbb{R}^{(n_t+1)\times H \times W}$ are the forward model and network predictions, respectively, for the hydraulic head field and $n_t$ concentration fields, and $\bar{\mathbf{y}}=1/N\sum_{i=1}^{N}\mathbf{y}^i$. A lower RMSE value and a $R^2$ score value approaching $1.0$ suggest better surrogate quality.

 \begin{table}[h!]
 \centering
 \caption{The input and output channels, and the batch size $M$ used in \texttt{Net}, \texttt{AR-Net}, and \texttt{AR-Net-WL}. $\mathbf{K}$ is the conductivity field, $\mathbf{S}_j$ denotes the image for the source term at the $j$-th time segment, $\mathbf{h}$ is the hydraulic head field, and $\mathbf{c}_i$ is the concentration field at the $j$-th time step.}
 \label{tab:net-diff}
 \begin{threeparttable}
 \begin{tabular}{lccc}
    \hline
    Network & Input channels & Output channels & Batch size $M$ \\
    \hline
    \texttt{Net}    & $\mathbf{K}$, $\{\mathbf{S}_j\}_{j=1}^{5}$ & $\mathbf{h}$, $\{\mathbf{c}_i\}_{i=1}^{7}$ & $30$       \\
    \texttt{AR-Net}       & $\mathbf{K}$, $\mathbf{S}_i$,  $\mathbf{c}_{i-1}$\tnote{$\dagger$} & $\mathbf{h}$, $\mathbf{c}_i$          & $200$     \\
    \texttt{AR-Net-WL}       & $\mathbf{K}$, $\mathbf{S}_i$,  $\mathbf{c}_{i-1}$\tnote{$\dagger$} & $\mathbf{h}$, $\mathbf{c}_i$          & $200$     \\
    \hline
\end{tabular}
\begin{tablenotes}\footnotesize
    \item[$\dagger$] Here $i=1,\ldots,7$ and $\mathbf c_0=\mathbf0$ is the initial concentration. For the last two time intervals when no contaminant is released, the input image for the source term is zero at all pixels, i.e., $\mathbf{S}_6$, $\mathbf{S}_7=\mathbf{0}$.
    \end{tablenotes}
\end{threeparttable}
\end{table}

\subsection{Network design}
 The differences between the autoregressive networks (i.e., \texttt{AR-Net} and \texttt{AR-Net-WL}) and the network without using the autoregressive strategy (i.e., \texttt{Net}) are summarized in Table~\ref{tab:net-diff}. They include the number of input and output channels, and the batch size $M$. In \texttt{Net}, there are six input channels $\mathbf x\in \mathbb R^{6\times 41\times 81}$ and eight output channels $\mathbf y\in \mathbb R^{8\times 41\times 81}$, i.e., $\mathbf x=\big(\mathbf K,\{\mathbf{S}_j\}_{j=1}^5\big)$ and $\mathbf y=\big(\mathbf h,\{\mathbf c_i\}_{i=1}^7\big)$, where $\mathbf{h}$ is the hydraulic head field, and $\mathbf{c}_i$ is the concentration field at the $j$-th time step. In \texttt{AR-Net} and \texttt{AR-Net-WL} in which the training data are reorganized using equation~(\ref{eq:data-reorg}), there are three input channels $\tilde{\mathbf x}\in \mathbb R^{3\times 41\times 81}$ and two output channels $\mathbf y\in \mathbb R^{2\times 41\times 81}$, i.e., $\tilde{\mathbf x}=(\mathbf K,\mathbf{S}_i, \mathbf c_{i-1})$ and $\mathbf y=(\mathbf h,\mathbf c_i)$, $i=1,\ldots,7$, where $\mathbf c_0=\mathbf 0$ is the initial concentration. For the last two time intervals when no contaminant is released, the input image for the source term is zero at all pixels, i.e., $\mathbf{S}_6$, $\mathbf{S}_7=\mathbf{0}$. In \texttt{Net}, one training sample contains $7$ concentration fields at seven time steps, while in \texttt{AR-Net} and \texttt{AR-Net-WL}, one training sample contains only $1$ concentration field after reorganizing the training data. Thus, the batch size used in \texttt{Net} is $M=30$ and in \texttt{AR-Net} and \texttt{AR-Net-WL} is $M=200$. 
 
 The design of the network configuration follows~\citet{zhu2018} and~\citet{mo2018} and the employed network architecture is shown in Table~\ref{tab:net-para}. The network is fully-convolutional consisting of $27$  layers. It contains three dense blocks with $L=5,\:10$ and $5$ internal layers, and a constant growth rate $R=40$. Three transition layers (one encoding layer and two decoding layers) are placed after each dense block. The number of initial feature maps after the first convolutional layer (the Conv in Figure~\ref{fig:DCEDN}) is $48$. The parameter values of the (transposed) convolution kernels (i.e., kernel size $k'$, stride $s$, and zero-padding $p$) used in the dense blocks and transition layers are also listed in Table~\ref{tab:net-para}, where their values are denoted by the number in the right, e.g., for $k'7s2p3$: $k'=7$, $s=2$, and $p=3$. Since the value of concentration is always non-negative, the softplus function which is written as
 \begin{linenomath*}
 \begin{equation}
     Softplus(x)=\frac{1}{\beta}\log{\big(1+\exp{(\beta x)}\big)},
 \end{equation}
 \end{linenomath*}
 is employed as the activation function of the output layer for the concentration to ensure non-negative predicted values, where $\beta=5$ is a shape parameter. In addition, the sigmoid function is used as the activation function of the output layer for the hydraulic head as its value varies between $0$ and $1$ in the case study. The initial learning rate $\varepsilon=0.005$ and weight decay $\omega_d=5\times10^{-5}$. We also use a learning rate scheduler which drops ten times on plateau during the training process. The network is trained for $200$ epochs.

 \begin{table}[h!]
 \centering
 \caption{Network architectures of \texttt{Net}, \texttt{AR-Net}, and \texttt{AR-Net-WL}. $N_{\rm{f}}$ denotes the number of output feature maps, $H_{\rm{f}}\times W_{\rm{f}}$ denotes the output feature map size, $L$ and $R$ are the number of internal layers and the growth rate of dense block, respectively.}
 \label{tab:net-para}
 \begin{threeparttable}
 \begin{tabular}{llccc}
    \hline
    \multirow{2}{*}{Layers}  & \multirow{2}{*}{Convolution kernels\tnote{$\dagger$}}                & \multicolumn{3}{c}{Feature maps ($N_{\rm{f}}\times H_{\rm{f}}\times W_{\rm{f}}$)}       \\
                         &                                                     & \texttt{Net} & \texttt{AR-Net} & \texttt{AR-Net-WL} \\
    \hline
    Input                    & -                                                   & $6\times 41\times 81$               & $3\times 41\times 81$              & $3\times 41\times 81$                 \\
    Convolution              & $k'7s2p3$                                           & \multicolumn{3}{c}{$48\times 21\times 41$}                                                                       \\
    Dense block 1 $(R40L5)$  & $k'3s1p1$\tnote{$\ddagger$}        & \multicolumn{3}{c}{$248\times 21\times 41$}                                                                      \\
    Encoding layer           & $k'1s1p0$, $k'3s2p1$\tnote{$\ast$} & \multicolumn{3}{c}{$124\times 11\times 21$}                                                                      \\
    Dense block 2 $(R40L10)$ & $k'3s1p1$                                           & \multicolumn{3}{c}{$524\times 11\times 21$}                                                                      \\
    Decoding layer 1         & $k'1s1p0$, $k'3s2p1$\tnote{$\ast$} & \multicolumn{3}{c}{$262\times 21\times 41$}                                                                      \\
    Dense block 3 $(R40L5)$  & $k'3s1p1$                                           & \multicolumn{3}{c}{$462\times 21\times 41$}                                                                      \\
    Decoding layer 2         & $k'1s1p0$, $k'5s2p2$\tnote{$\ast$} & $8\times 41\times 81$               & $2\times 41\times 81$              & $2\times 41\times 81$                       \\
    \hline                         
    \end{tabular}
    \begin{tablenotes}\footnotesize
        \item[$\dagger$]The values of kernel size $k'$, stride $s$, and zero-padding $p$ are denoted by the number in the right. The relationship between output $H_{\rm{fy}}$ $(W_{\rm{fy}})$ and input feature size $H_{\rm{fx}}$ $(W_{\rm{fx}})$ of the convolution is given in equation~(\ref{eq:featureSize}) and of the transposed convolution is $H_{\rm{fy}}=s(H_{\rm{fx}}-1)+k'-2p$~\citep{dumoulin2016}. 
        \item[$\ddagger$]The convolution kernels are the same for all convolutional layers within the dense blocks.
        \item[$\ast$] The first and second convolution kernels correspond to the first and second (transposed) convolutions of the encoding/decoding layers, respectively. 
    \end{tablenotes}
    \end{threeparttable}
\end{table}

\section{Results and Discussion}\label{sec:results}
 In this section, we first assess the network's performance in approximating the contaminant transport system with a time-varying contaminant source strength. After that, the inversion results obtained from our deep autoregressive network-based ILUES are compared to those obtained from the ILUES without surrogate modeling. 
 
\subsection{Approximation Accuracy Assessment}\label{sec:approx-accuracy}
\begin{figure}[h!]
 \centering
    \includegraphics[width=0.6\textwidth]{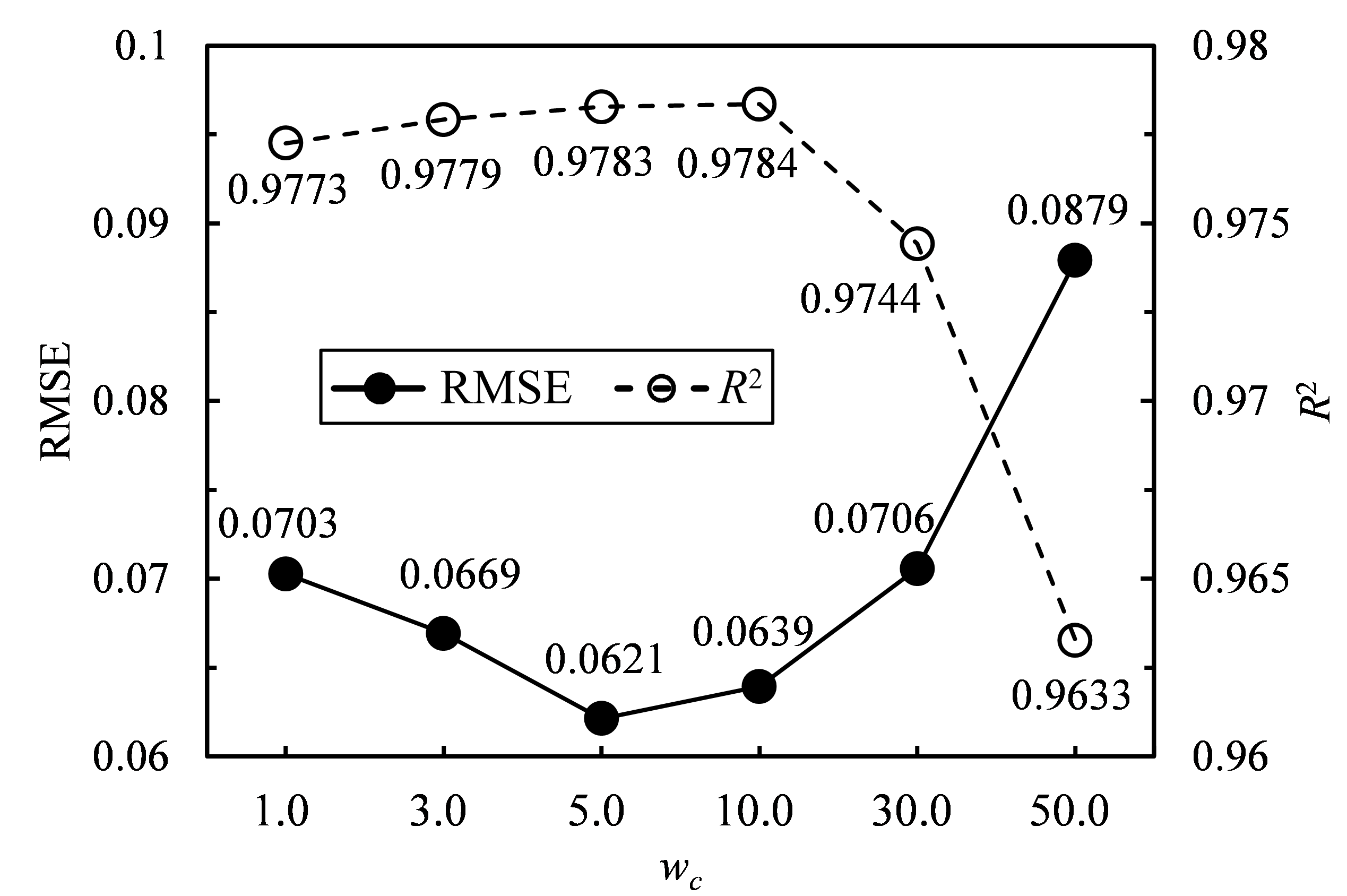}
     \caption{RMSEs and $R^2$ scores of the $\texttt{AR-Net-WL}$ evaluated on the test dataset with six different loss weights $w_c$ used in equation~(\ref{eq:weighted-loss}). The networks are trained using $N=400$ model evaluations.}
    \label{fig:r2-rmse-weight}
 \end{figure}

 To determine a good value of the weight $w_c$ in the weighted loss (equation~(\ref{eq:weighted-loss})) used in $\texttt{AR-Net-WL}$, six $\texttt{AR-Net-WL}$ networks were trained using $400$ model evaluations with different weight values varying from $1.0$ to $50.0$ to test the effect of $w_c$ on the network's performance. Figure~\ref{fig:r2-rmse-weight} shows the RMSEs and $R^2$ scores of the six $\texttt{AR-Net-WL}$s evaluated on the test dataset. It is observed that the $\texttt{AR-Net-WL}$s with small $w_c$ values ($1.0$, $3.0$, $5.0$, and $10.0$) perform better than those with large $w_c$ values ($30.0$ and $50.0$). A large $w_c$ value promotes the training process to heavily focus on local refinement of the approximation for the large-gradient concentration region near the source release location, leading to a deterioration in the global approximation accuracy. Among the six trained $\texttt{AR-Net-WL}$ networks, the one with $w_c=5.0$ performs well achieving a low RMSE value and a high $R^2$ score. Therefore, $w_c=5.0$ is selected for the remaining of our numerical studies.
 
\begin{figure}
    \centering
    \includegraphics[width=\textwidth]{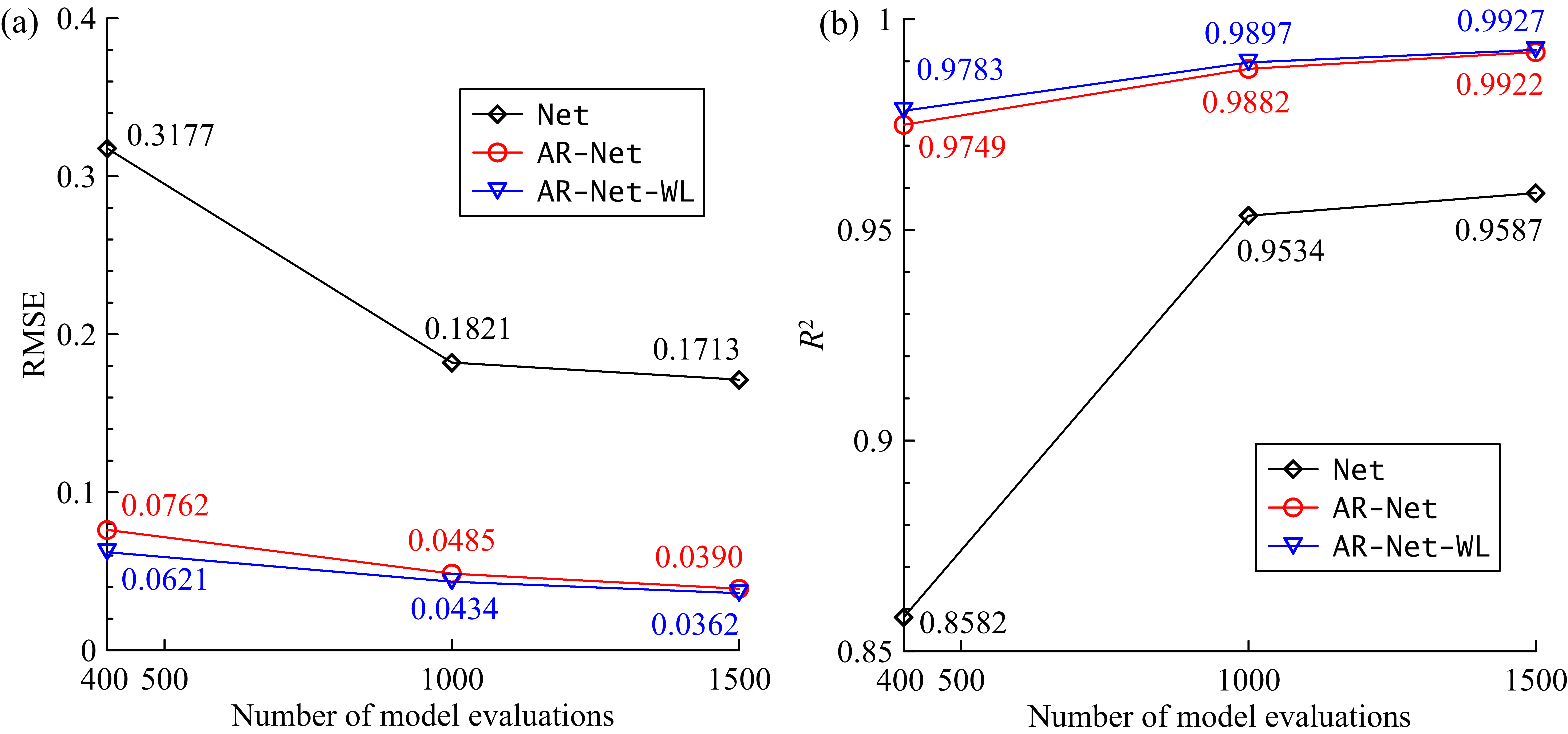}
    \caption{Comparison of the (a) RMSEs and (b) $R^2$ scores of \texttt{Net}, \texttt{AR-Net}, and \texttt{AR-Net-WL} evaluated on the test dataset. The networks are trained using three different training sample sizes of $N=400$, $1000$, and $1500$.}
    \label{fig:rmse-r2}
 \end{figure}

 Next, we compare the performance of the $\texttt{Net}$, $\texttt{AR-Net}$, and $\texttt{AR-Net-WL}$ networks trained using three different sample sizes ($N=400,1000,1500$). The networks are trained on a NVIDIA GeForce GTX $1080$ Ti X GPU. The training time of $\texttt{Net}$ is about $6-18$ minutes for training $200$ epochs when the training sample size varies from $400$ to $1500$, and that of $\texttt{AR-Net}$ and $\texttt{AR-Net-WL}$ is about $32-110$ minutes. The increase in training time of $\texttt{AR-Net}$ and $\texttt{AR-Net-WL}$ is due to the number of training samples being increased to $n_t$ times  the number of forward model runs after reorganizing the data. The networks' RMSEs and $R^2$ scores evaluated on the test dataset are shown in Figure~\ref{fig:rmse-r2}. For the solute transport system with a time-varying source release strength, the figure clearly demonstrates that the autoregressive networks $\texttt{AR-Net}$ and $\texttt{AR-Net-WL}$ achieve much lower RMSEs and much higher $R^2$ values than the $\texttt{Net}$ network. For example, with $N=400$ forward model evaluations for network training, $\texttt{Net}$ only achieves a RMSE value of $0.3177$ and $R^2$ value of $0.8581$, much lower than that of $\texttt{AR-Net}$ and $\texttt{AR-Net-WL}$ which achieve RMSE values of $0.0762$ and $0.0621$, and $R^2$ values of $0.9749$ and $0.9783$, respectively. With more training data available, the difference between the RMSEs and $R^2$ scores of the $\texttt{Net}$ and $\texttt{AR-Net}$/$\texttt{AR-Net-WL}$ decreases. With $1500$ forward evaluations for surrogate construction, $\texttt{AR-Net}$ and $\texttt{AR-Net-WL}$ can provide rather accurate approximations with $R^2$ scores of $0.9922$ and $0.9927$, respectively, for a problem with high-dimensional input and output fields. The results indicate that the autoregressive strategy employed in the $\texttt{AR-Net}$ and $\texttt{AR-Net-WL}$ networks substantially improves the network's performance in approximating the solute transport system with a time-varying source term especially when the available training data are limited.
 
 \begin{figure}[h!]
 \centering
 \includegraphics[width=\textwidth]{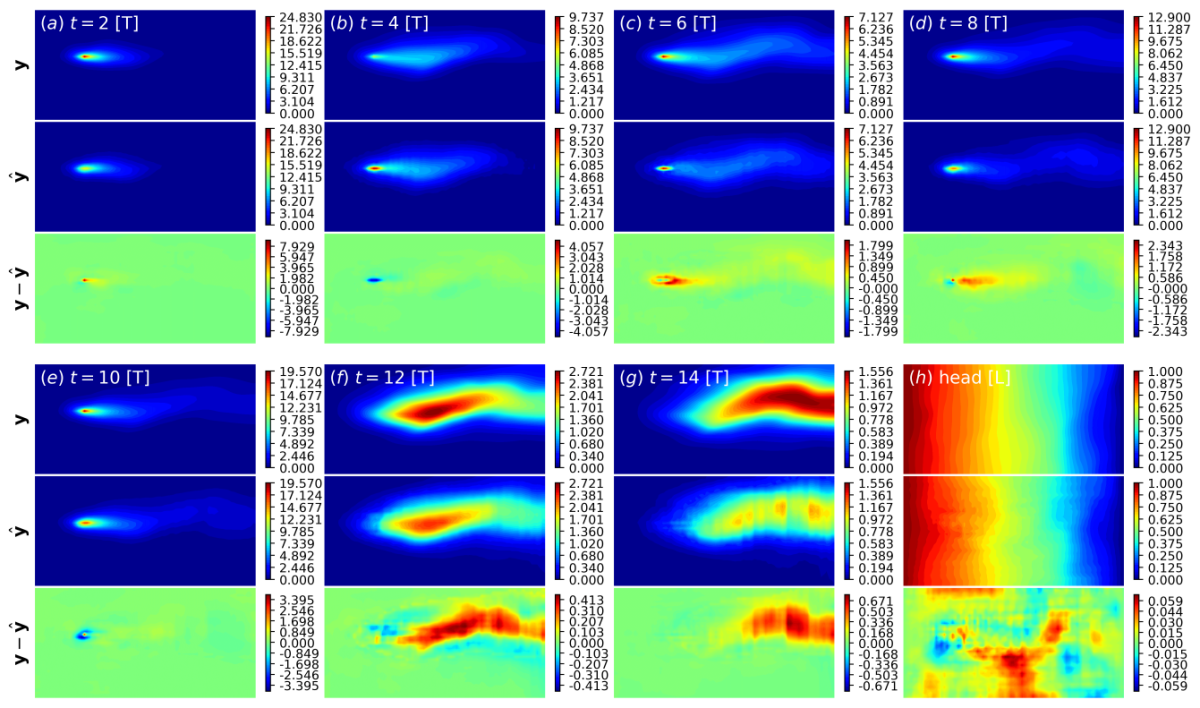}
     \caption{Snapshots of the concentration fields at time $t=2,4,\ldots,14$ [T] (a-g) and hydraulic head field (h) of a random test sample predicted by the forward model $(\mathbf{y})$ and the \texttt{Net} network $(\hat{\mathbf{y}})$ trained using $N=400$ model evaluations. $(\mathbf{y}-\hat{\mathbf{y}})$ denotes the difference between the predictions of the forward model and network.}
 \label{fig:pred-Net-woTDR-N400}
 \end{figure}
 
\begin{figure}[h!]
 \centering
 \includegraphics[width=\textwidth]{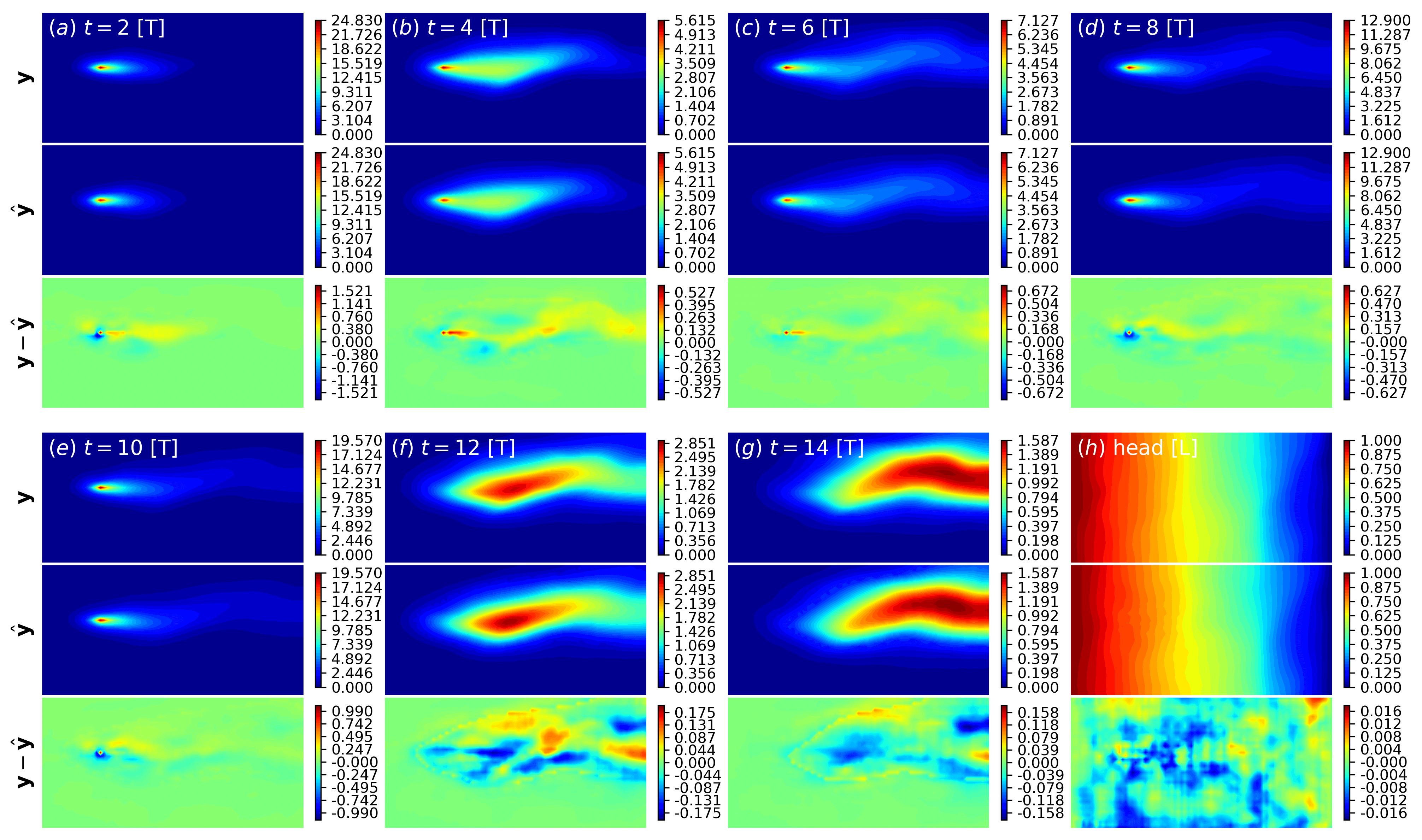}
     \caption{Snapshots of the concentration fields at time $t=2,4,\ldots,14$ [T] (a-g) and hydraulic head field (h) of a random test sample predicted by the forward model $(\mathbf{y})$ and the \texttt{AR-Net} network $(\hat{\mathbf{y}})$ trained using $N=400$ model evaluations. $(\mathbf{y}-\hat{\mathbf{y}})$ denotes the difference between the predictions of the forward model and network.}
 \label{fig:pred-Net-wTDR-N400}
 \end{figure}

\begin{figure}[h!]
 \centering
 \includegraphics[width=\textwidth]{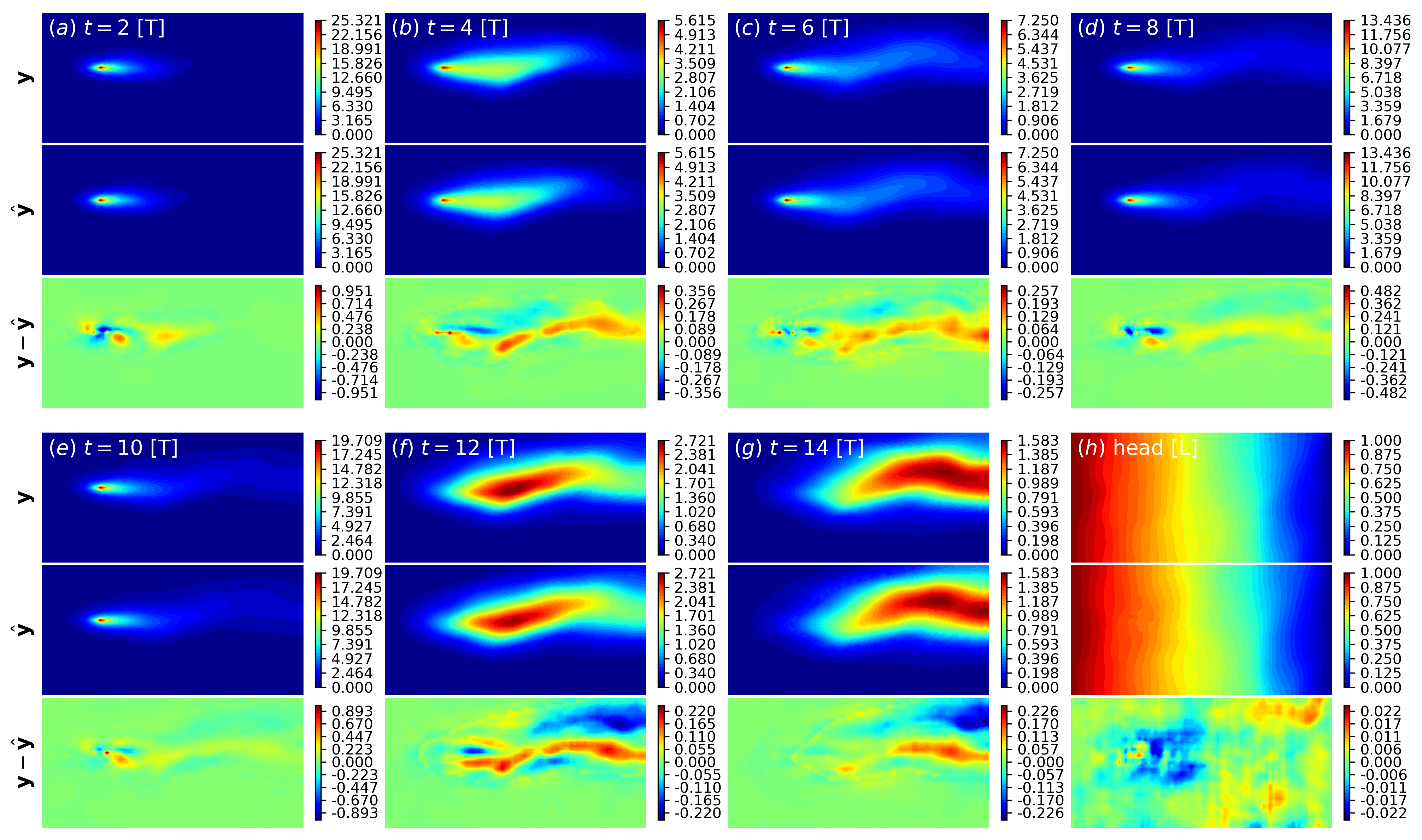}
     \caption{Snapshots of the concentration fields at time $t=2,4,\ldots,14$ [T] (a-g) and hydraulic head field (h) of a random test sample predicted by the forward model $(\mathbf{y})$ and the \texttt{AR-Net-WL} network $(\hat{\mathbf{y}})$ trained using $N=400$ model evaluations. $(\mathbf{y}-\hat{\mathbf{y}})$ denotes the difference between the predictions of the forward model and network.}
 \label{fig:pred-Net-wTDR+WL-N400}
 \end{figure}

 The performance of the networks is further compared in Figures~\ref{fig:pred-Net-woTDR-N400}, \ref{fig:pred-Net-wTDR-N400}, and \ref{fig:pred-Net-wTDR+WL-N400}, which respectively show the predictions of \texttt{Net}, \texttt{AR-Net}, and \texttt{AR-Net-WL} using $N=400$ training samples for the concentration fields at $t=2,4,\ldots,14$ [T] and the hydraulic head field of a random test sample. For comparison, we also show the forward model predictions in each plot. Figure~\ref{fig:pred-Net-woTDR-N400} shows that the predictions of \texttt{Net} have relatively large approximation errors for both the concentration and hydraulic head fields. For the concentration fields at the five time steps at $t=2,4,\ldots,10$ [T] when the contaminant is released, the approximation errors in the large-gradient region near the source release location are large relative to other regions. Figure~\ref{fig:pred-Net-wTDR-N400} shows that \texttt{AR-Net} significantly reduces the approximation errors compared to those of \texttt{Net}. However, the approximation errors in the five concentration fields at $t=2,4,\ldots,10$ [T] near the source release location are still relatively large. In \texttt{AR-Net-WL},  we assign an additional weight of $w_c=5.0$ in the loss to the nine pixels around the source release location in the five concentration fields (equation~(\ref{eq:weighted-loss})). Figure~\ref{fig:pred-Net-wTDR+WL-N400} indicates that the weighted loss in \texttt{AR-Net-WL} brings in an additional improvement in the characterization of this sharp-concentration feature without deteriorating the overall approximation accuracy. This improvement can also be demonstrated by Figure~\ref{fig:rmse-r2} which shows that \texttt{AR-Net-WL} achieves slightly lower RMSEs and higher $R^2$ scores than \texttt{AR-Net} when trained using the same three training datasets. 
 
\begin{figure}[h!]
 \centering
    \includegraphics[width=\textwidth]{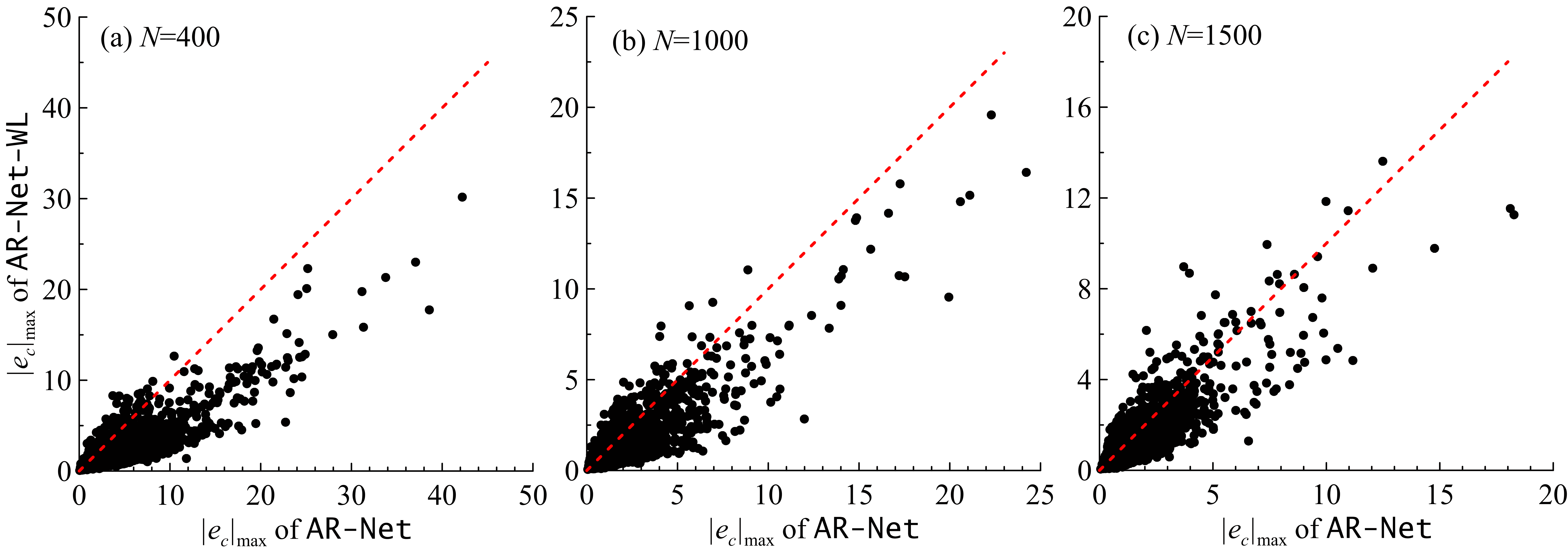}
     \caption{Point-wise comparison of the maximum of the absolute prediction errors $|e_c|_{\text{max}}$ of the \texttt{AR-Net} and \texttt{AR-Net-WL} in each field of the $2500$ test concentration fields at $t=2,4,\ldots,10$~[T] (i.e., the concentration fields at the $n'_t=5$ time steps in the $N_{\rm test}=500$ test samples). The \texttt{AR-Net} and \texttt{AR-Net-WL} networks are trained using three different training sample sizes of $N=400$, $1000$, and $1500$. The dash line denotes the $(1:1)$ line. Statistics of $|e_c|_{\text{max}}$ are summarized in Table~\ref{tab:comp-cmax}.}
    \label{fig:comp-cmax}
 \end{figure}

 As can be seen from Figures~\ref{fig:pred-Net-woTDR-N400}-\ref{fig:pred-Net-wTDR+WL-N400}, the maximum value of the absolute approximation errors, $|e_c|_{\max}=\max\big(|\mathbf y-\hat{\mathbf y}|\big)$, in each field of the $2500$ concentration fields at $t=2,4,\ldots,10$ [T] (i.e., the concentration fields at the $n'_t=5$ time steps in the $N_{\rm test}=500$ test samples) usually occurs in the sharp-concentration region near the source release location. The improvement of \texttt{AR-Net-WL} compared to \texttt{AR-Net} in characterizing this sharp-concentration feature is further illustrated in Figure~\ref{fig:comp-cmax}. It depicts a point-wise comparison of the $2500$ predictive values of $|e_c|_{\max}$ of the two networks trained with $N=400, 1000$, and $1500$ in the $2500$ test concentration fields. An $(1:1)$ line is added in each plot to facilitate the comparison. A data point below (above) the $(1:1)$ line indicates that \texttt{AR-Net} has a larger (smaller) $|e_c|_{\max}$ value in a test concentration field than \texttt{AR-Net-WL}. It is observed that more data points are below the $(1:1)$ line in all three cases with $N=400, 1000$, and $1500$, indicating \texttt{AR-Net-WL}'s  overall better performance in approximating the sharp-concentration feature near the source release location. This can also be demonstrated in Table~\ref{tab:comp-cmax} which summarizes the mean and standard deviation of the $2500$ values of $|e_c|_{\max}$ obtained by the two networks. The table shows that the \texttt{AR-Net-WL} achieves lower means and standard deviations of $|e_c|_{\max}$ than \texttt{AR-Net} in all three cases with $N=400, 1000$, and $1500$. 
 
  \begin{table}[h!]
    \centering
    \caption{Comparison of the mean and standard deviation (std) of the maximum of the absolute prediction errors $|e_c|_{\text{max}}$ of \texttt{AR-Net} and \texttt{AR-Net-WL} in each field of $2500$ test concentration fields at $t=2,4,\ldots,10$~[T] (i.e., the concentration fields at the $n'_t=5$ time steps in the $N_{\rm test}=500$ test samples). $N$ denotes the training sample size.}
    \label{tab:comp-cmax}
    \begin{tabular}{lcccc}
    \hline
    \multirow{2}{*}{N} & \multicolumn{2}{c}{Mean of $|e_c|_{\rm{max}}$} & \multicolumn{2}{c}{Std of $|e_c|_{\rm{max}}$} \\
                  & \texttt{AR-Net}             & \texttt{AR-Net-WL}             & \texttt{AR-Net}             & \texttt{AR-Net-WL}            \\
    \hline
    400  & 3.4930                 & 1.9432                    & 4.0837                 & 2.3279                   \\
    1000 & 1.8300                 & 1.2930                    & 2.1042                 & 1.5650                   \\
    1500 & 1.4002                 & 1.1140                    & 1.5095                 & 1.2727      \\
    \hline
    \end{tabular}
 \end{table}

\begin{figure}[h!]
 \centering
 \includegraphics[width=\textwidth]{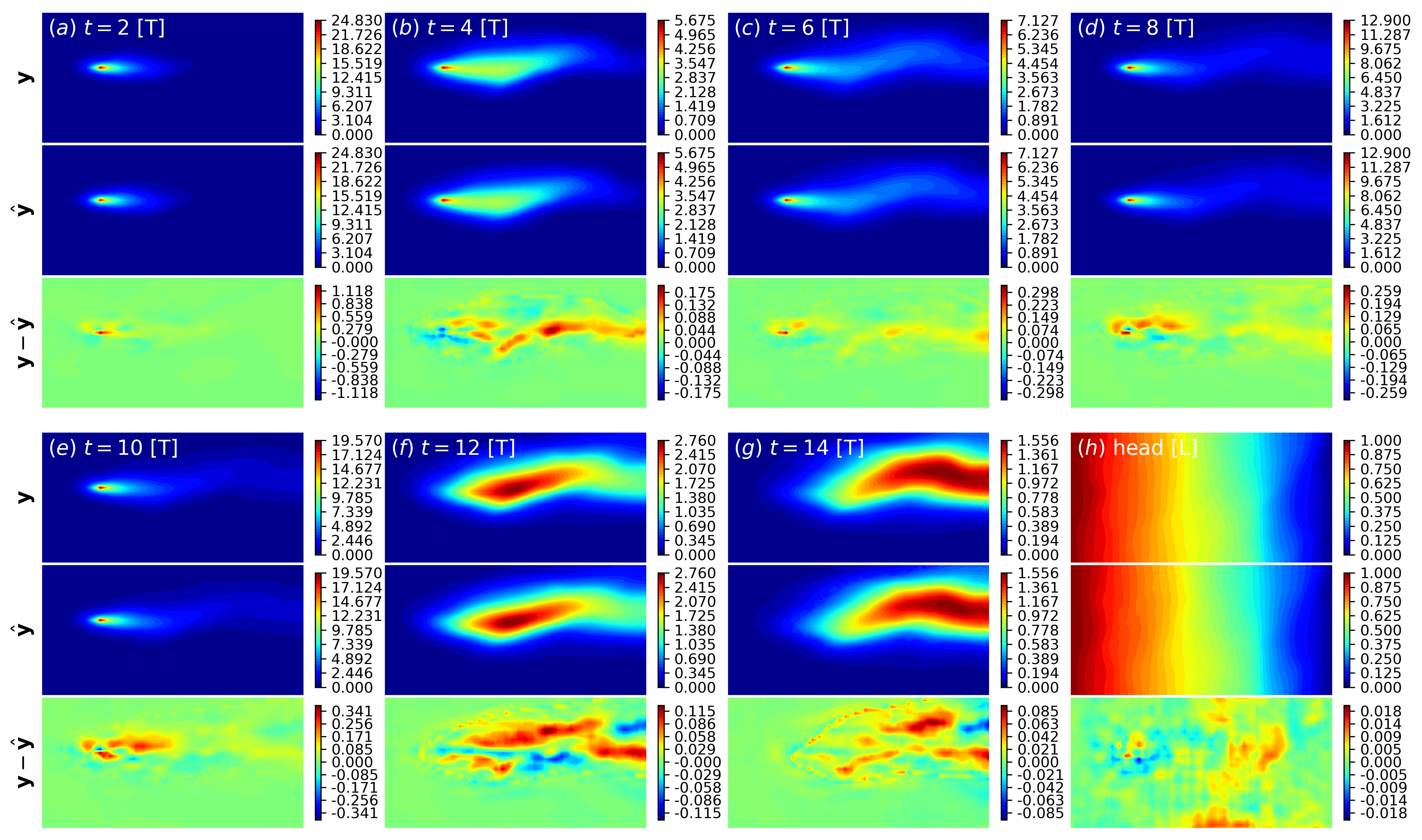}
     \caption{Snapshots of the concentration fields at time $t=2,4,\ldots,14$ [T] (a-g) and hydraulic head field (h) of a random test sample predicted by the forward model $(\mathbf{y})$ and the \texttt{AR-Net-WL} network $(\hat{\mathbf{y}})$ trained using $N=1500$ model evaluations. $(\mathbf{y}-\hat{\mathbf{y}})$ denotes the difference between the predictions of the forward model and network.}
 \label{fig:pred-Net-wTDR+WL-N1500}
 \end{figure}

\begin{figure}[h!]
 \centering
 \includegraphics[width=\textwidth]{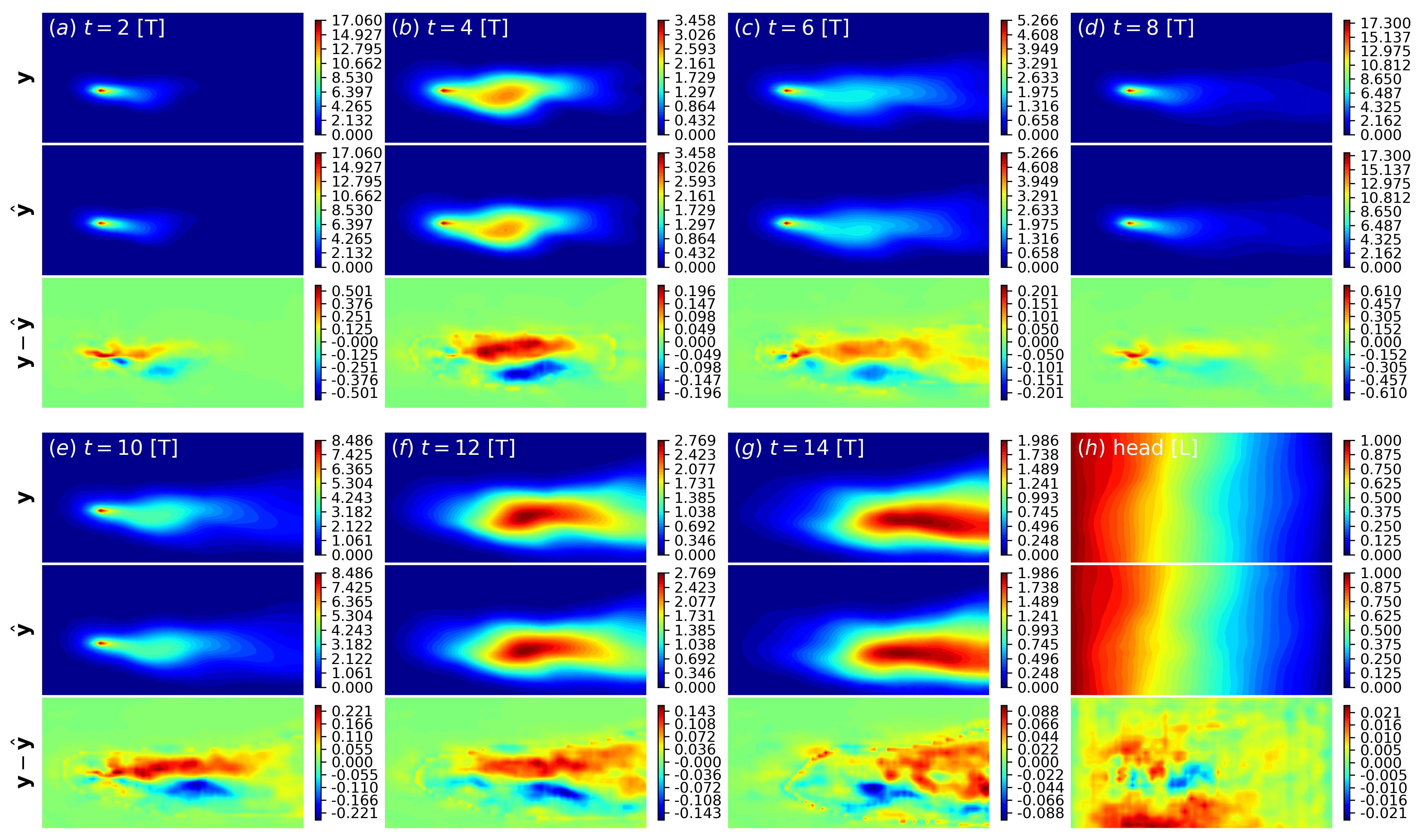}
     \caption{Snapshots of the concentration fields at time $t=2,4,\ldots,14$ [T] (a-g) and hydraulic head field (h) of the reference log-conductivity field shown in Figure~\ref{fig:conceptual_model} and source parameters listed in Table~\ref{tab:prior} predicted by the forward model $(\mathbf{y})$ and the \texttt{AR-Net-WL} network $(\hat{\mathbf{y}})$ trained using $N=1500$ model evaluations. $(\mathbf{y}-\hat{\mathbf{y}})$ denotes the difference between the predictions of the forward model and network.}
 \label{fig:pred-refInput-Net-wTDR+WL-N1500}
 \end{figure}

 The results presented above indicate a good performance of \texttt{AR-Net-WL} in comparison to \texttt{Net} and \texttt{AR-Net} in approximating the contaminant transport system with a time-varying source term. Thus the \texttt{AR-Net-WL} network with $N=1500$ is chosen as the surrogate model that will be used to substitute the forward model in the inverse problem in section~\ref{sec:inverse} to alleviate the computational burden. Its predictions for the same output fields as those shown in Figures~\ref{fig:pred-Net-woTDR-N400}-\ref{fig:pred-Net-wTDR+WL-N400} are given in Figure~\ref{fig:pred-Net-wTDR+WL-N1500}. An improved approximation accuracy can be seen after increasing the training sample size from $400$ to $1500$. The predictions for the output fields of the reference log-conductivity field shown in Figure~\ref{fig:conceptual_model} and the source parameters listed in Table~\ref{tab:prior} are depicted in Figure~\ref{fig:pred-refInput-Net-wTDR+WL-N1500}, which again suggests an accurate surrogate prediction. 
 
 The results above illustrate the good performance of the employed deep dense convolutional encoder-decoder network architecture~\citep{mo2018,zhu2018} in accurately and efficiently approximating a solute transport system with high-dimensional inputs and outputs. This is attributed to the image-to-image regression strategy and the densely connected convolutional network architecture. The employment of the autoregressive strategy representing the time-varying process substantially improves the surrogate (i.e., \texttt{AR-Net} and \texttt{AR-Net-WL}) quality without bringing additional computational costs in terms of  forward model evaluations. In addition, the weighted loss in \texttt{AR-Net-WL}, which assigns an additional weight to the large-gradient concentrations near the source release location, further improves the surrogate predictive capability.

 \subsection{Inversion Results}\label{sec:inverse}

 In surrogate-based ILUES, the \texttt{AR-Net-WL} network trained with $N=1500$ forward model evaluations is used as a full-replacement of the forward model, which means no additional forward model runs are required in the inversion process. To assess the accuracy of the obtained inversion results and the computational efficiency of the surrogate method, the ILUES algorithm using forward model without surrogate modeling (referred to as the original ILUES hereinafter) is also performed to compute a reference result. In both surrogate-based and original ILUES, an ensemble size of $N_e=6000$ is chosen for the high-dimensional inverse problem, a local ensemble factor of $\alpha=0.1$ suggested in~\citet{zhang2018} is used, and $N_{\text{iter}}=20$ iterations are performed. Thus, $(20+1)\times 6,000=126,000$ forward model executions (i.e., $1$ prior ensemble and $20$ updated ensembles) are performed in the original ILUES algorithm to obtain the reference solution.

 We first assess the performance and convergence of the original ILUES   that is used as the reference method. To this end, we calculate the convergence of the sum of squared weighted residuals (SSWR) measure which quantifies the mismatch between the predicted outputs and measurements, as given by
 \begin{linenomath*}
 \begin{equation}
     SSWR=\sum_{i=1}^{N_d}\Big(\frac{f_i(\bm m)-d_i}{\sigma_i}\Big)^2,
 \end{equation}
 \end{linenomath*}
 where $\bm\sigma=[\sigma_1,\ldots,\sigma_{N_d}]$ are the standard deviation of the measurement errors, and $\big\{f_i(\bm m)\big\}_{i=1}^{N_d}$ and $\big\{d_i\big\}_{i=1}^{N_d}$ are the model predictions and measurements, respectively. 
 
  \begin{figure}[h]
 \centering
 \includegraphics[width=0.55\textwidth]{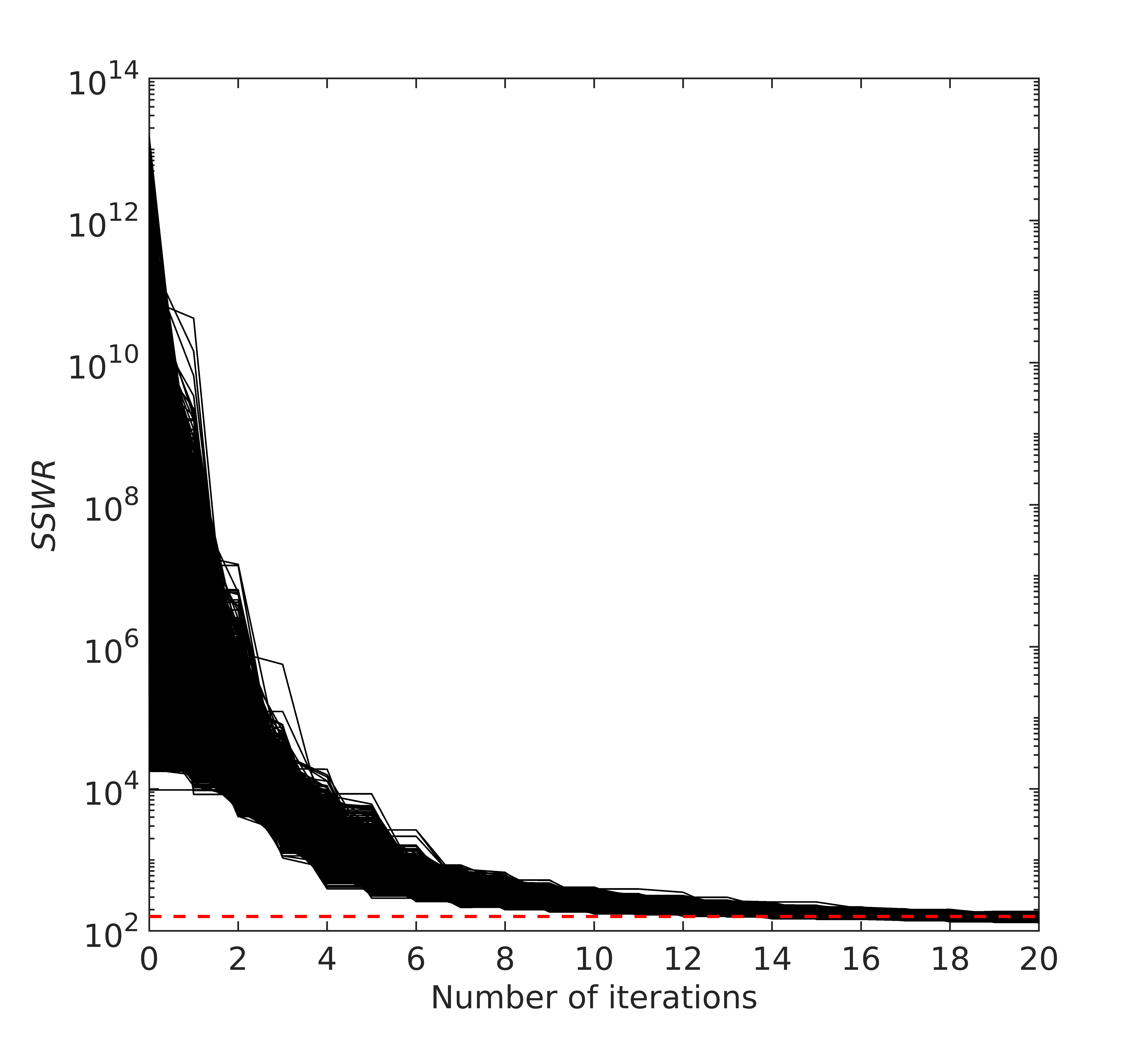}
     \caption{Convergence of the $SSWR$s of the $6000$ samples in the ensemble of the original ILUES with forward model as the number of iterations increases. The reference $SSWR$ value is denoted by the dashed line. }
 \label{fig:sswr}
 \end{figure}
 
\begin{figure}[h!]
 \centering
 \includegraphics[width=0.8\textwidth]{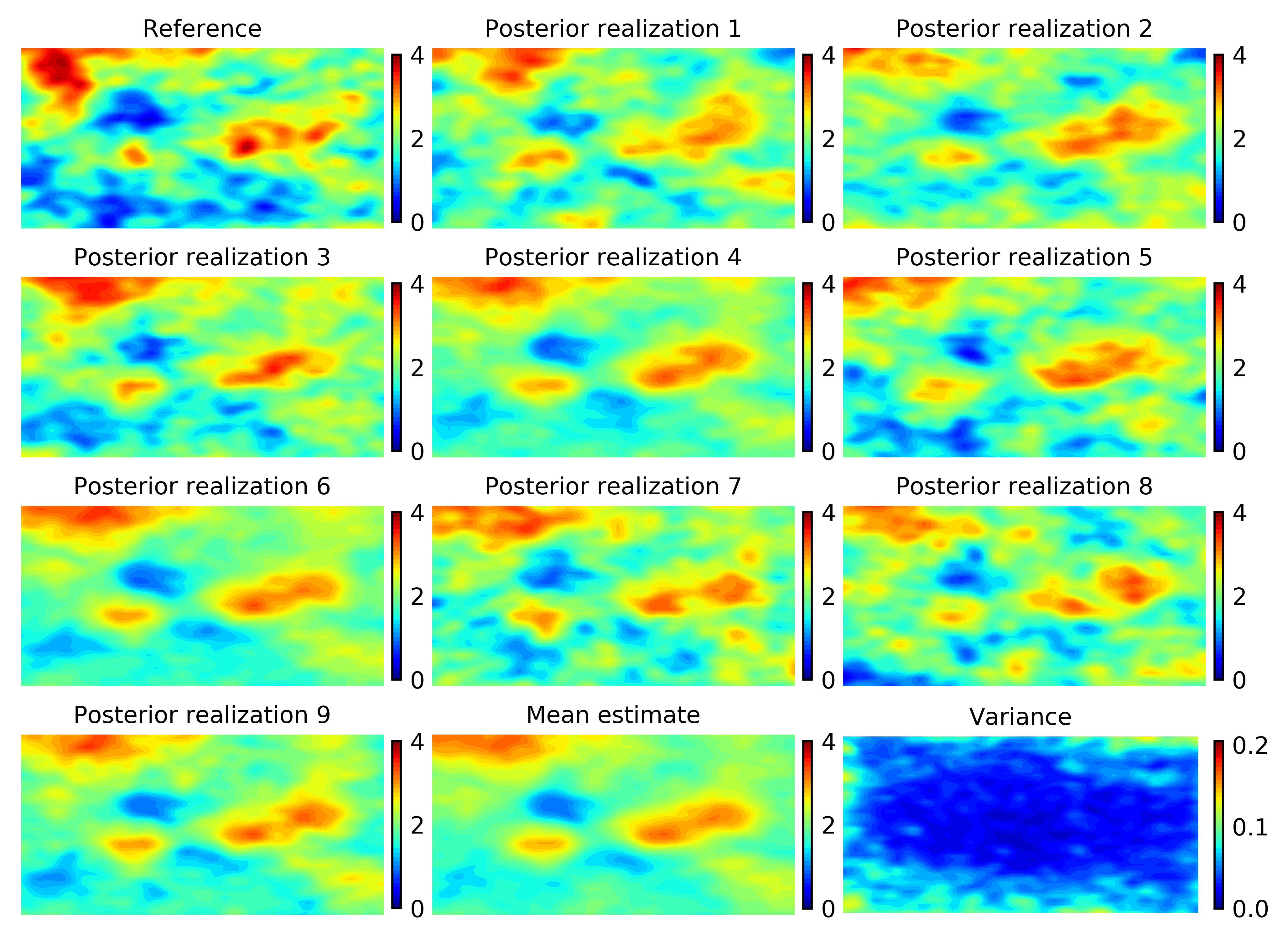}
     \caption{Reference log-conductivity field, nine posterior realizations, mean and variance fields obtained from the original ILUES algorithm with $126,000$ forward model runs.}
 \label{fig:Kfield-ILUES}
 \end{figure}

\begin{figure}[h!]
 \centering
 \includegraphics[width=\textwidth]{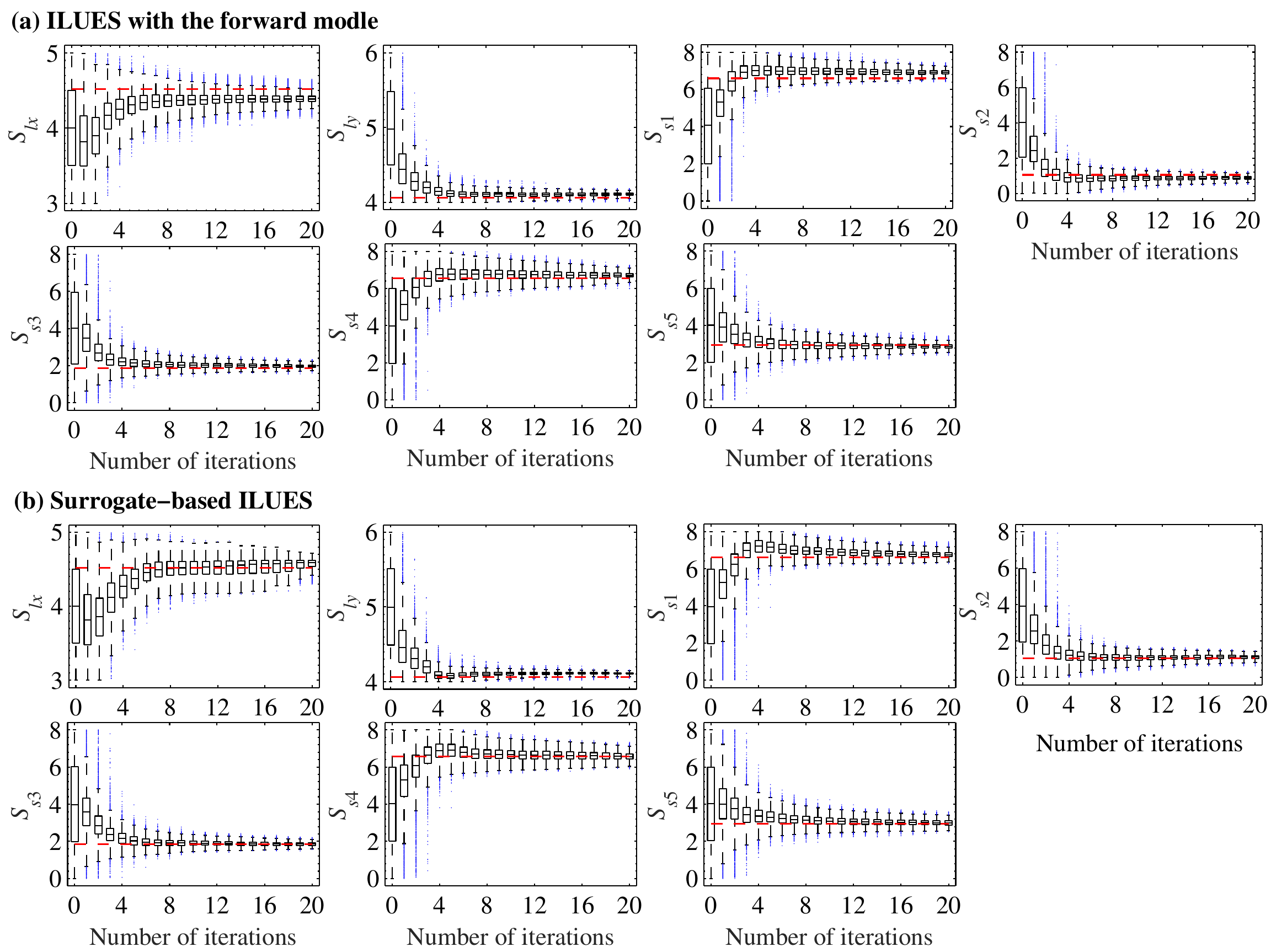}
     \caption{Box plots of the source location ($S_{lx},S_{ly}$) and strength $\{S_{sj}\}_{j=1}^5$ at different iterations obtained by (a) the original ILUES algorithm with the forward model and (b) the surrogate-based ILUES method. The dashed horizontal red line denote the reference value. The numbers of forward model runs needed by the original and surrogate-based ILUES are $126,000$ and $1500$, respectively.}
 \label{fig:boxplot}
 \end{figure}

 Figure~\ref{fig:sswr} shows the convergence of the $SSWR$ of each sample in the ensemble of the original ILUES algorithm as the number of iterations increases. It can be seen that the $SSWR$s of the $6,000$ samples in the ensemble converge after about $16$ iterations. Therefore, the posterior samples at iteration $20$ are used to compute the reference solutions. Nine posterior realizations and the mean and variance estimates of the log-conductivity field obtained from the original ILUES are shown in Figure~\ref{fig:Kfield-ILUES}. The reference log-conductivity field is also shown in the plot to facilitate the comparison. It is observed that the original ILUES can accurately capture the high-conductivity and low-conductivity regions of the reference field.  Due to the relatively sparse observations (using $168$ output observations to infer $686$ uncertain input parameters), the local conductivity in the posterior realizations may not match the reference accurately. Figure~\ref{fig:boxplot}a depicts the box plots of the $7$ source parameters (the location $(S_x,S_y)$ and strength $\{S_{sj}\}_{j=1}^5$) versus the number of iterations. The figure indicates that these parameters can be accurately identified by the ILUES using the original forward model.
 
\begin{figure}[h!]
 \centering
 \includegraphics[width=0.8\textwidth]{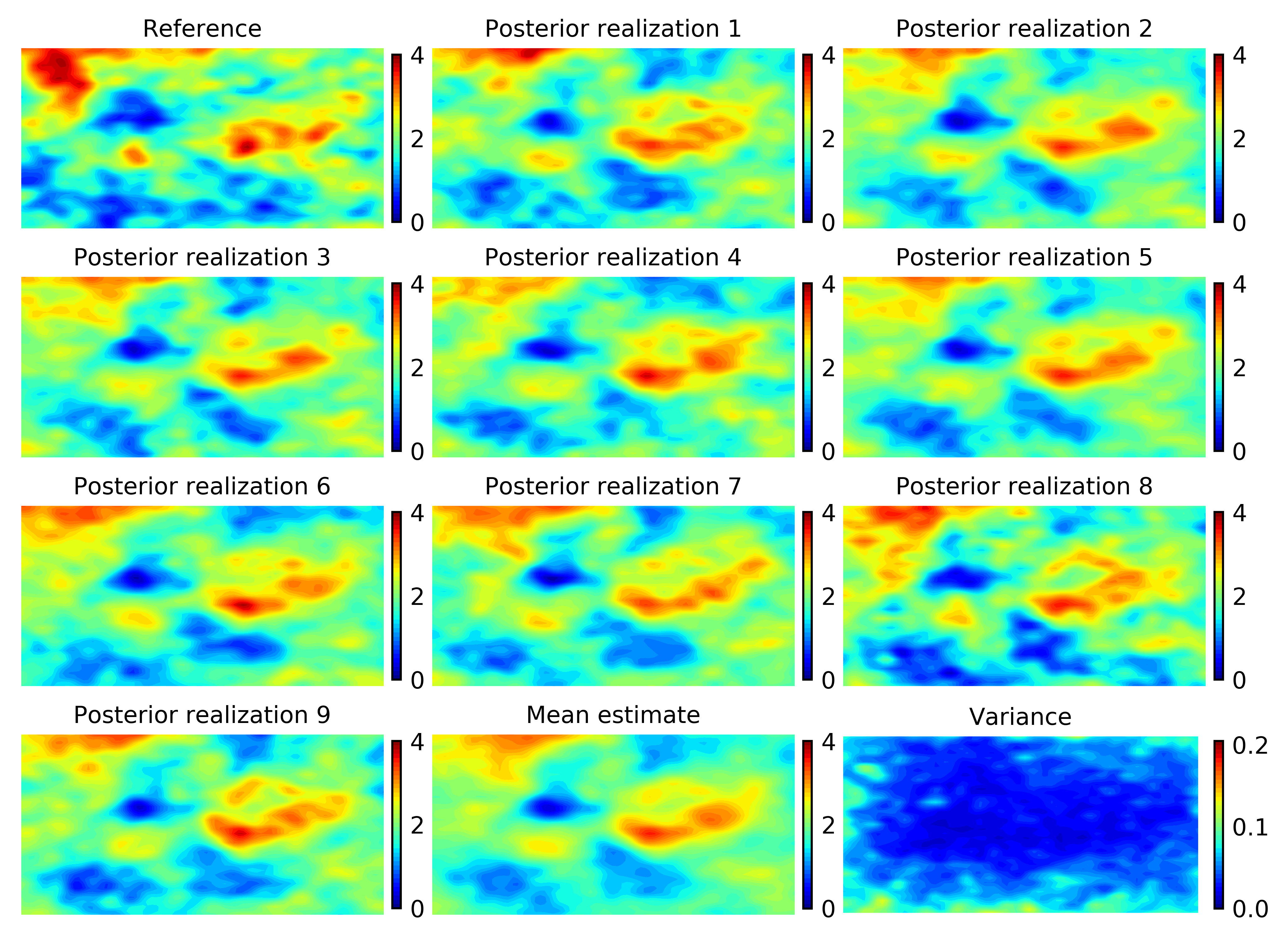}
     \caption{Reference log-conductivity field, nine posterior realizations, mean and variance fields obtained from the surrogate-based ILUES method. The surrogate model is trained using $1500$ forward model runs.}
 \label{fig:Kfield-DNN}
 \end{figure}

 The above results indicate that the ILUES algorithm works well for the $686$-dimensional  inverse problem considered but requires a large number (i.e., $126,000$) of forward model runs. Next we shows the results of our deep autoregressive neural network based ILUES with only $1500$ forward model runs for network training. Figure~\ref{fig:Kfield-DNN} illustrate nine posterior realizations and the mean and variance estimates of the log-conductivity field obtained from the surrogate-based ILUES algorithm. The box plots of the $7$ source parameters with the number of itertions are depicted in Figure~\ref{fig:boxplot}b. Similarly, the surrogate-based ILUES method accurately captures the high-conductivity and low-conductivity regions of the reference field and successfully identifies the source parameters. It is worth noting that in the surrogate-based ILUES the fast-to-evaluate surrogate is used and no additional forward model executions are required. Thus, the comparably accurate inversion results suggest that the surrogate-based method substantially improves the computational efficiency of the inversion for this case. 
 
 \begin{figure}[h!]
 \centering
 \includegraphics[width=\textwidth]{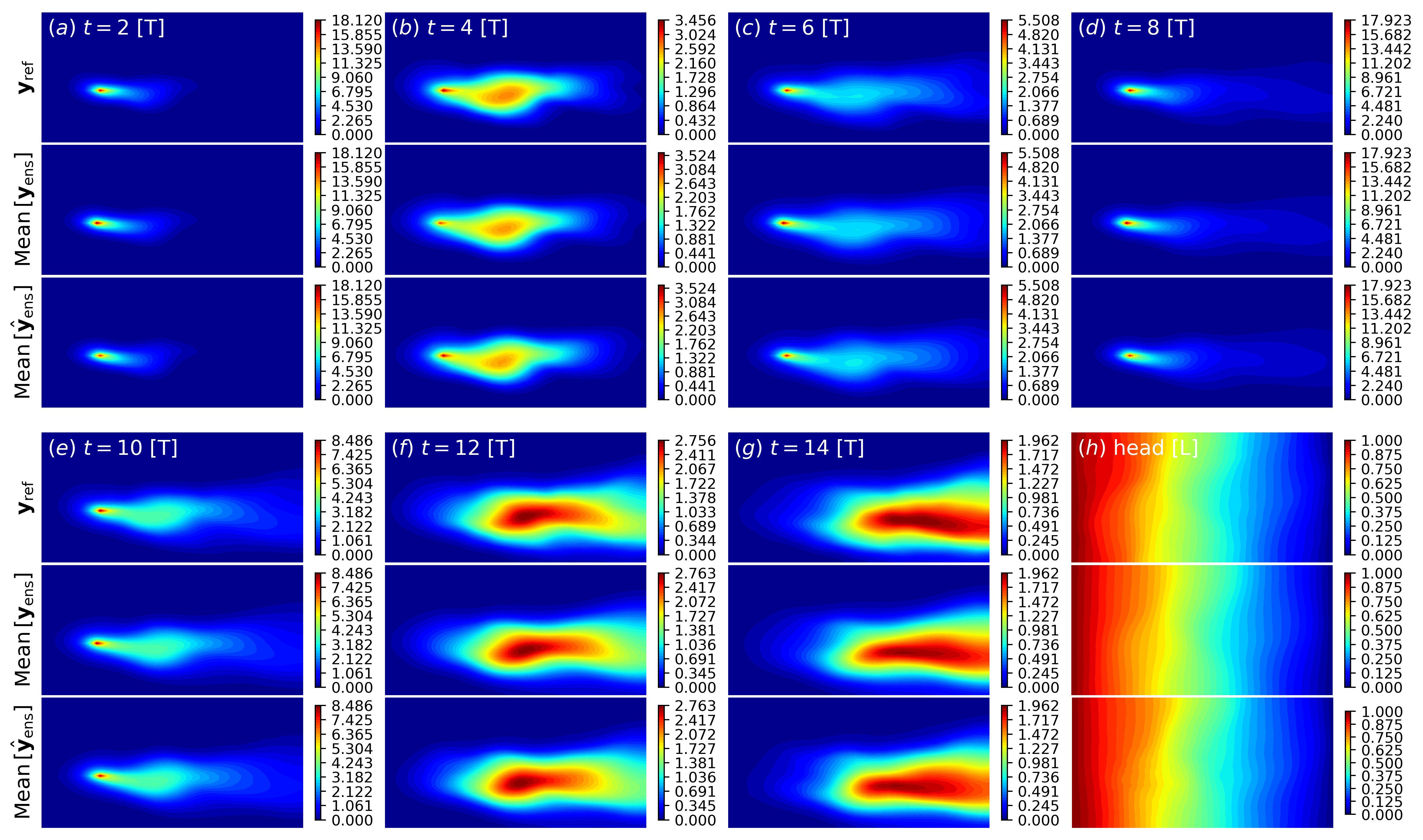}
     \caption{Ensemble mean of the output concentration fields at $t=[2,4,\ldots,14]$ [T] and hydraulic head field obtained from the original ($\text{Mean}\,[\mathbf y_{\text{ens}}]$) and surrogate-based ($\text{Mean}\,[\hat{\mathbf y}_{\text{ens}}]$) ILUES. The output fields in the reference model ($\mathbf y_{\text{ref}}$) are also shown for comparison. The numbers of forward model runs needed by the original and surrogate-based ILUES are $126,000$ and $1500$, respectively.}
 \label{fig:ens-mean}
 \end{figure}
 
\begin{figure}[h!]
 \centering
 \includegraphics[width=\textwidth]{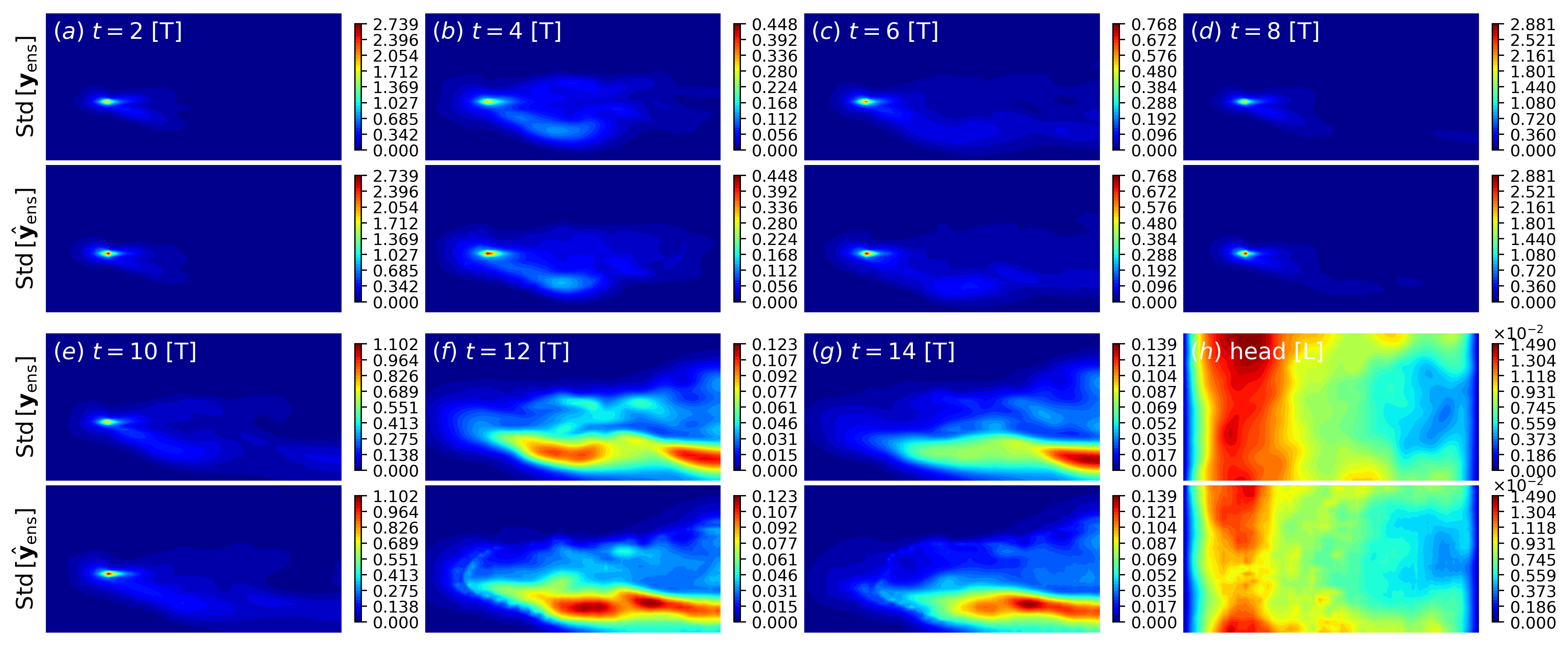}
     \caption{Ensemble standard deviation (std) of the output concentration fields at $t=[2,4,\ldots,14]$ [T] and hydraulic head field obtained from the original ($\text{Std}\,[\mathbf y_{\text{ens}}]$) and surrogate-based ($\text{Std}\,[\hat{\mathbf y}_{\text{ens}}]$) ILUES. The numbers of forward model runs needed by the original and surrogate-based ILUES are $126,000$ and $1500$, respectively.}
 \label{fig:ens-std}
 \end{figure}

 Based on the estimated input parameter ensemble, the predictive uncertainty associated with the output concentration and hydraulic head can be obtained. Figures~\ref{fig:ens-mean} and~\ref{fig:ens-std} depict the ensemble mean and standard deviation fields obtained from the ILUES with the forward model and surrogate model. The output fields in the reference model are also shown in Figure~\ref{fig:ens-mean} for comparison. Notice that statistics of the surrogate-based ILUES are computed using the outputs predicted by the surrogate model instead of the forward model. From the two figures, we can see that the mean and standard deviation fields obtained by the ILUES with the original forward model and surrogate model are very similar and their mean fields visually are almost identical to the reference fields. The standard deviation of concentration in regions near the source release location  (Figure~\ref{fig:ens-std}a-e) is large relative to other regions due to large concentration variability near the source. In addition, when the contaminant plume reaches the lower-right part of the domain at  $t=12$ and $14$ [T] (Figure~\ref{fig:ens-mean}f-g), the standard deviation of the concentration in this area is also relatively large because no observations from this region are available (Figure~\ref{fig:refK}). The results indicate that, with a small number of $1500$ model evaluations for surrogate construction, the surrogate-based method is capable of efficiently obtaining accurate estimations of the high-dimensional uncertain input parameters and of the predictive uncertainty associated with the high-dimensional output fields. 
 
\section{Conclusions}\label{sec:conclusion}

 In this study, we propose a deep autoregressive neural network based iterative ensemble smoother to efficiently solve the high-dimensional inverse problem of  identification of the groundwater contaminant source simultaneously with the hydraulic conductivity field. The highly-heterogeneous conductivity leads to high-input dimensionality, posing a great challenge for constructing accurate surrogate models efficiently. To tackle the curse of dimensionality, a deep neural network (DNN) based on a convolutional encoder-decoder network architecture is employed. In the network, the high-dimensional input and output fields are treated as images. The surrogate modeling is then transformed to an image-to-image regression task. To efficiently obtain an accurate surrogate model for the contaminant transport model with a time-varying source term, we represent the time-varying process using an autoregressive model, in which the time-dependent output at the previous time step ($y_{i-1}$) is treated as input to predict the current output ($y_i$), i.e., $y_i=f(x_i,y_{i-1})$, where $x$ is the uncertain model input considered. The use of the autoregressive strategy enables the network to clearly capture the complex relationship between the time-varying model inputs and outputs. To further overcome the data-intensive training of DNNs, we apply a densely connected convolutional network architecture which introduces connections between nonadjacent layers to reduce the network parameters and to enhance the information flow through the network. 
 
 To improve the network's approximation accuracy in capturing the strongly nonlinear feature of concentration in the region near the source release location, an additional loss term to the $L_1$ loss function for network training is assigned for the concentration values in this region. The iterative local updating ensemble smoother (ILUES) method proposed in~\citet{zhang2018} is used as the inversion framework.
 
 The performance of the proposed method is demonstrated using a synthetic $686$-dimensional inverse problem of the joint identification of the contaminant source and highly-heterogeneous conductivity field. The results indicate that our deep autoregressive neural network can provide an accurate approximation for the mapping between high-dimensional inputs (the hydraulic conductivity field and source term) and outputs (the hydraulic head field and time serials concentration fields) of a dynamical contaminant transport system using only a small number of forward model evaluations. The autoregressive strategy substantially improves the network's performance in approximating the time-varying process. The additional loss term assigned to concentrations near the source release location brings in an additional improvement in approximating the large-concentration gradients. The application of the surrogate-based ILUES method in solving the inverse problem shows that it can achieve comparably accurate inversion results and predictive uncertainty estimations to those obtained by the original ILUES algorithm without surrogate modeling but requiring much fewer forward model evaluations. 
 
Although not consider in the case study, it is straightforward to combine our surrogate method with design of experiments~\citep [e.g.,][]{huan2013} to efficiently select the most informative observations for performing inference of the unknown parameters and thus reduce the cost of measurement collections~\citep{Rajabi2018}. The proposed DNN method is non-intrusive and its potential use as a surrogate method to be combined with other inversion or uncertainty analysis methods for many complex systems beyond groundwater solute transport remains to be explored.
 
\acknowledgments
The work of N.Z. was supported  from the Defense Advanced Research Projects Agency (DARPA) under the Physics of Artificial Intelligence (PAI) program (contract HR$00111890034$). Additional computing resources were provided by the University of Notre Dame's Center for Research Computing (CRC). S.M. acknowledges the China Scholarship Council for financially supporting his study at the Center for Informatics and Computational Science (CICS) at the University of Notre Dame. The work of S.M., X.S. and J.W. was supported by the National Natural Science Foundation of China (No. U$1503282$ and $41672229$).  The Python codes and data used will be made available at \url{https://github.com/cics-nd/cnn-inversion} upon publication of this manuscript.


\end{document}